\PassOptionsToPackage{table}{xcolor}
\documentclass[]{fairmeta}

\microtypesetup{expansion=false}

\usepackage{amsmath}
\usepackage{amsfonts}
\usepackage{amssymb}
\usepackage{pifont}
\usepackage{siunitx}
\usepackage{enumitem}
\usepackage{adjustbox}
\usepackage{makecell}
\usepackage{wrapfig}
\usepackage{threeparttable}
\usepackage{tabularx}
\usepackage{array}
\usepackage[ruled,vlined,linesnumbered]{algorithm2e}
\usepackage{eccvabbrv}

\newcolumntype{Y}{>{\raggedright\arraybackslash}X}
\SetAlgoNoEnd
\SetAlgoNlRelativeSize{-1}
\DontPrintSemicolon
\crefname{algocf}{algorithm}{algorithms}
\Crefname{algocf}{Algorithm}{Algorithms}

\newcommand{\cmark}{\ding{51}}
\newcommand{\xmark}{\ding{55}}
\newcommand{\answerTODO}[1][]{\textcolor{red}{\bfseries [TODO]}}
\newcommand{\justificationTODO}[1][]{\textcolor{red}{\bfseries [TODO]}}

\title{VL-LN Bench: Towards Long-horizon Goal-oriented Navigation with Active Dialogs}

\author[1,2,*]{Wensi Huang}
\author[2,3,*]{Shaohao Zhu}
\author[2,4]{Meng Wei}
\author[2,5]{Siqi Zhang}
\author[3]{Jinming Xu}
\author[4]{Xihui Liu}
\author[2,\dagger]{Hanqing Wang}
\author[2,\dagger]{Tai Wang}
\author[1,\dagger]{Feng Zhao}
\author[2]{Jiangmiao Pang}
\affiliation[1]{University of Science and Technology of China}
\affiliation[2]{Shanghai AI Laboratory}
\affiliation[3]{Zhejiang University}
\affiliation[4]{The University of Hong Kong}
\affiliation[5]{Tongji University}
\contribution[*]{Equal Contribution}
\contribution[\dagger]{Corresponding Author}
\metadata[Keywords]{Goal-oriented Navigation; Active Interaction; Navigation Benchmark}

\abstract{%
In most existing embodied navigation tasks, instructions are well-defined and unambiguous, such as instruction following and object searching. Under this idealized setting, agents are required solely to produce effective navigation outputs conditioned on vision and language (VL) inputs. Real-world instructions, however, are often underspecified and require interaction to resolve ambiguity and infer user intent. To bridge this gap, we propose Interactive Instance Goal Navigation (IIGN), which extends Instance Goal Navigation (IGN) by allowing agents to freely consult an oracle in natural language while searching for a specific instance. IIGN requires agents to produce both Language and Navigation (LN) outputs, enabling interaction while moving in the environment. To support this task, we introduce VL-LN Bench, a benchmark with an automated data collection pipeline and over 41k collected long-horizon dialog-augmented trajectories for training, alongside an automatic evaluation protocol paired with a dedicated oracle for answering agent queries. Experiments reveal two core bottlenecks of IIGN: long-horizon exploration and fine-grained grounding of textual information to the correct instance among same-category distractors. Although active dialog partially alleviates these challenges, current models still lag far behind human performance. Further ablations validate the value of the data generated by our pipeline and show that the proposed oracle provides scalable assistance comparable to human support, proving VL-LN Bench as a practical testbed for dialog-enabled embodied navigation. Code and dataset can be found at \href{https://0309hws.github.io/VL-LN.github.io/}{https://0309hws.github.io/VL-LN.github.io/}.
}

\begin{document}

\maketitle

\begin{figure}[t]
  \centering
  \includegraphics[width=\linewidth]{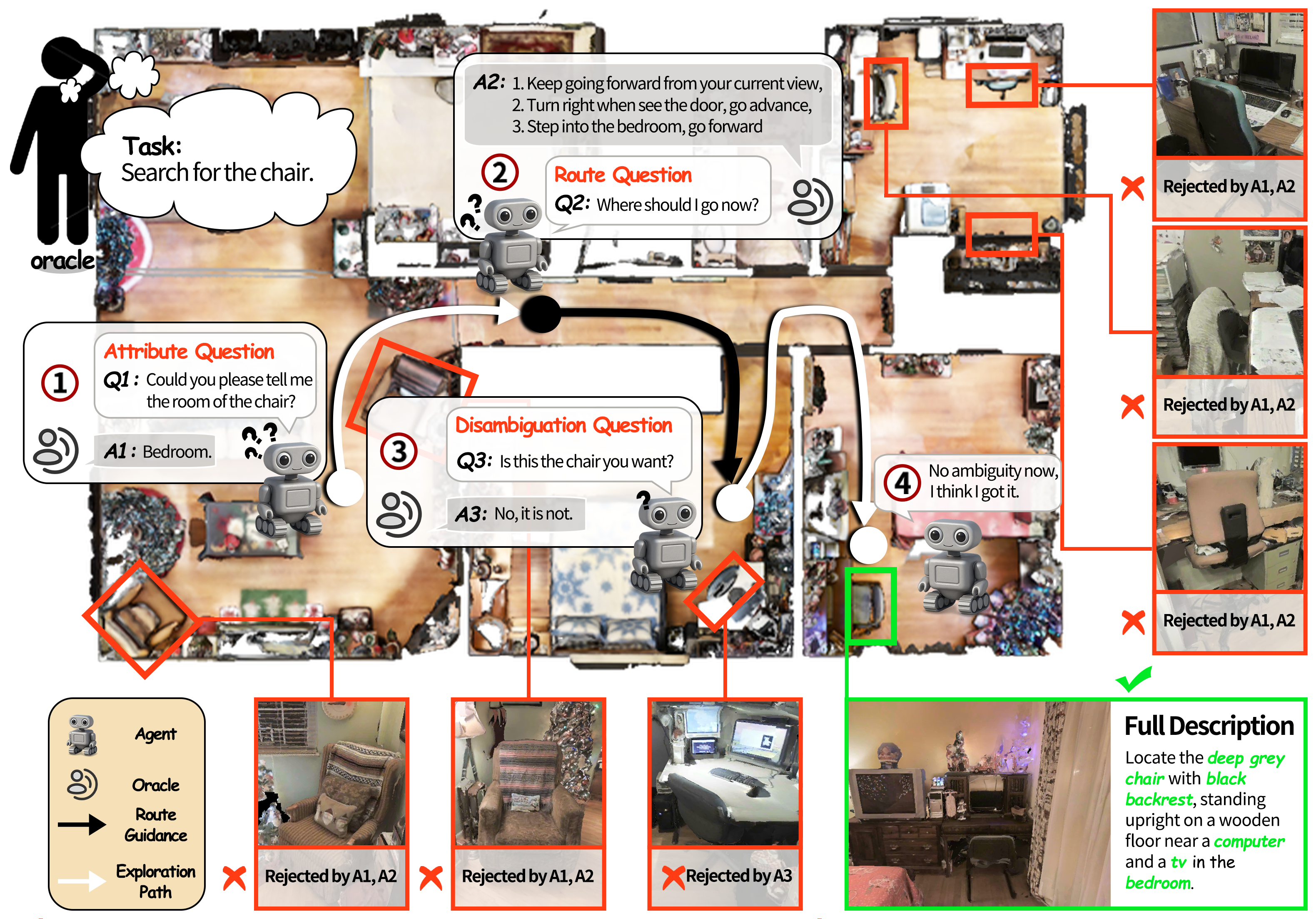}
  \caption{IIGN requires the agent to actively engage in free-form, open-ended natural-language dialog during navigation to clarify underspecified goals and improve navigation efficiency. In this example, the oracle (top left) provides an initially underspecified instruction, ``Search for the chair.'', requiring the agent to locate a specific instance of the target category (chair) in an unfamiliar environment. Through three effective dialog turns---including attribute query, route guidance, and disambiguation question---the agent progressively eliminates incorrect candidates (red box) and ultimately identifies the target instance (green box). The description shown in the bottom right serves as the IGN instruction, which can specify the target chair among multiple distractor chairs.}
  \label{fig:teaser}
\end{figure}

\section{Introduction}
A practical navigation agent must handle vague tasks by both planning effectively and resolving ambiguities~\cite{25,26,27}. Active dialog offers a natural solution~\cite{41,majumdar2022zson,12}, allowing the agent to clarify underspecified instructions~\cite{majumdar2023findthis} and obtain cues for efficient navigation. To study this ability, we propose Interactive Instance Goal Navigation (IIGN; as shown in ~\cref{fig:teaser}), which extends Instance-goal Navigation (IGN) \cite{ION}. In IIGN, the agent receives a basic ObjectNav \cite{objectnav,yokoyama2024hm3d} instruction (\eg,``Search for the \textless category\textgreater''), which is insufficient to uniquely identify the target instance, and must consult an oracle via dialog. Since IIGN inherently combines long-horizon exploration with target disambiguation, it provides an ideal testbed for evaluating whether an agent can actively acquire useful information and translate it into effective navigation decisions.

To effectively investigate and address the proposed IIGN task, this paper aims to build an Agent–Oracle interaction benchmark for dialog-enabled navigation agent training and evaluation. Prior work \cite{thomason2020vision} collected human–human dialogs in navigation to evaluate whether agents can understand and follow instructions. However, such efforts do not assess an agent’s ability to proactively ask targeted questions. More recent approaches attempt to enable agents to generate questions, but they either limit the task to small-scale, room-level settings \cite{gao2022dialfred} or focus narrowly on instance descriptions and provide limited support for long-horizon exploration \cite{taioli2024collaborative}. Moreover, existing methods generally lack large-scale training datasets, which constrains the development of agents capable of active exploration and informative questioning.

In contrast to prior work, we first address the lack of training data, which is essential for teaching agents when to ask questions and how to leverage dialog-acquired information for navigation. Since manual data collection is costly and labor-intensive, we develop an automated data collection pipeline with three steps: (1) aggregating region-level room and instance attributes from MMScan~\cite{mmscan} into unified house-level annotations; (2) pairing target instances with feasible initial positions to construct episodes; and (3) collecting dialog-rich trajectories using a navigator with frontier-based exploration (FBE)~\cite{yamauchi1997frontier} and a GPT-based oracle. Using this pipeline, we curate the first large-scale training dataset for IIGN, containing approximately 41K dialog-augmented trajectories. 

Designing a comprehensive Agent--Oracle interaction mechanism is also a key aspect of the benchmark, since it supports both scalable data collection and online evaluation. Although the space of possible queries is open-ended, we find that three question types are particularly important for IIGN: \emph{attribute}, \emph{disambiguation}, and \emph{route}, as illustrated in~\cref{fig:teaser}. Attribute and disambiguation questions together enable the agent to identify the target instance, even among visually similar candidates, while route questions support long-horizon exploration in large-scale environments. To enable such interaction, we implement a scalable GPT-based~\cite{gpt4o} oracle that supports these core question types while remaining flexible to other queries. Building on our data collection pipeline and agent--oracle interaction support, we introduce VL-LN Bench, a unified benchmark for training and evaluating dialog-enabled agents in instance-goal navigation.

Extensive experiments on VL-LN Bench provide three main insights. First, compared with ObjectNav, both IGN and IIGN are intrinsically more challenging, not only because they require longer-horizon exploration, but more importantly because agents struggle to align textual information from either the initial instruction or dialog with the correct instance among same-category distractors. Second, dialog consistently improves performance by enhancing exploration efficiency and providing discriminative cues for instance disambiguation, yet current models still lag far behind human performance. Third, our ablation studies confirm the value of the data generated by our automatic pipeline and show that the proposed oracle offers scalable assistance comparable to human support. Moreover, we demonstrate the real-world feasibility of our unified VLLN model through deployment on a quadruped robot.
\FloatBarrier
\section{Related Works}
\textbf{Goal-oriented navigation.} Goal-oriented navigation~\cite{yin2025unigoal,qi2020reverie} requires an agent to locate a target object in an unknown environment~\cite{ieong2025multimodal}. In text-guided settings, this problem is commonly divided into Object-goal Navigation (ObjectNav)~\cite{chen2023not,anderson2018evaluation,objectnav,cai2024bridging} and Instance-goal Navigation (IGN)~\cite{ION,zheng2025learning}. Most prior work~\cite{campari2020exploiting,majumdar2022zson,xie2023implicit,yadav2023offline,vlfm,ye2021auxiliary,yu2023l3mvn} focuses on ObjectNav\cite{objectnav}, where the agent is given an instruction such as ``search for computer'' and only needs to reach any instance of the target category. In real-world scenarios, however, users often care about a particular~\cite{zhu2021soon,wang2026user} or personalized~\cite{barsellotti2024personalized} target rather than an arbitrary category instance. To better capture this setting, IGN~\cite{barsellotti2024personalized,sun2024prioritized} requires the agent to locate a specific instance based on a detailed textual description. Recent benchmarks~\cite{ION,barsellotti2024personalized} identify two central challenges in this setting: disambiguating the target instance and recovering after the agent moves in the wrong direction. Moreover, real environments often contain many visually similar instances, such as multiple chairs around the same table, while textual descriptions alone are often insufficient for reliable identification~\cite{taioli2024collaborative}. To address these limitations, we introduce interactive instance-goal navigation (IIGN), where the agent can proactively ask questions to refine the goal specification and obtain targeted guidance during navigation. This setting reduces the need for detailed instructions, making it more practical for daily use while improving discrimination between highly similar instances. 

\begin{table}[!ht]
\centering
\small
\setlength{\tabcolsep}{3pt}
\renewcommand{\arraystretch}{1.12}
\caption{Comparison with existing interactive navigation benchmarks.}
\label{tab:dataset_comparison}
\resizebox{\linewidth}{!}{%
\begin{tabular}{@{}lcccccccccc@{}}
\toprule
 & \multicolumn{5}{c}{\textbf{Training Dataset Attributes}} &
   \multicolumn{1}{c}{\textbf{Instruction}} &
   \multicolumn{4}{c}{\textbf{Evaluation Support}} \\
\cmidrule(lr){2-6}\cmidrule(lr){7-7}\cmidrule(lr){8-11}
\textbf{Dataset} &
\textbf{\#Trajectories} &
\makecell{\textbf{\#Dialogs}\\\textbf{(Q/A)}} &
\makecell{\textbf{Scene}\\\textbf{Scale}} &
\makecell{\textbf{Episode}\\\textbf{Length}} &
\makecell{\textbf{Annotation}\\\textbf{Source}} &
\makecell{\textbf{F. / P.}} &
\textbf{Oracle} &
\makecell{\textbf{Attr.}} &
\makecell{\textbf{Disamb.}} &
\makecell{\textbf{Route}} \\
\midrule
\rowcolor{gray!10}
CVDN~\cite{thomason2020vision}      & 7,000  & 2,050  & house & 25 (steps) & Human     & \xmark~/ \cmark & \xmark & -      & -      & - \\
DialFRED~\cite{gao2022dialfred}  & -     & 53,000 & room  & -          & Human     & \xmark~/ \cmark & \cmark & \cmark & \xmark & \cmark (dir.) \\
\rowcolor{gray!10}
CoIN~\cite{taioli2024collaborative}      & -  & -     & - & -    & -         & \xmark~/ \cmark & \cmark & \cmark & \xmark & \xmark \\
VL-LN (Ours)     & 41,891 & 95,559 & house & \makecell{22.5 (m)} & GPT+Rules & \cmark~/ \cmark & \cmark & \cmark & \cmark & \cmark (4m traj.) \\
\bottomrule
\end{tabular}}

\begin{tablenotes}[flushleft]
\scriptsize
\item Columns: \textbf{F.} (full instruction), \textbf{P.} (partial instruction), \textbf{Attr.} (attribute questions), \textbf{Disamb.} (disambiguation questions), \textbf{Dir.} (oracle-provided direction to the target), and \textbf{4 m Traj.} (oracle-provided natural-language trajectory for the next \(4\,\mathrm{m}\)). \textbf{VL-LN Bench} provides large-scale training data, an open-source generation pipeline, and more comprehensive evaluation for long-horizon interactive instance-goal navigation.
\end{tablenotes}
\end{table}
\FloatBarrier

\noindent\textbf{Interactive Navigation Benchmarks.} Interactive navigation benchmarks~\cite{han2025dialnav} investigate how an embodied agent follows ambiguous instructions and interacts with an oracle to obtain additional information for completing a navigation task. To enable such evaluation, prior benchmarks have introduced different forms of oracle support. DialFRED~\cite{gao2022dialfred} provides an oracle that returns target attributes and directional guidance, but only in small, single-room scenes. CoIN~\cite{taioli2024collaborative} employs a VLM-based oracle that can answer attribute-related questions with access to an image of the target. However, these designs are insufficient for instance-goal navigation in realistic house-scale environments, where agents must both explore over long horizons and disambiguate among visually similar instances. In addition, most existing benchmarks emphasize evaluation while providing limited support for large-scale data collection for policy learning~\cite{gao2022dialfred,taioli2024collaborative}. Although CVDN~\cite{thomason2020vision} collected human--human dialogs for object search, its scale is still insufficient for training agents that must both explore actively and converse effectively. To address these limitations, we introduce VL-LN Bench, an instance-goal navigation benchmark with a dialog-augmented data collection pipeline and a house-scale oracle that supports attribute, route, and disambiguation queries, enabling both efficient policy training and comprehensive evaluation. A detailed comparison of existing benchmarks is provided in \cref{tab:dataset_comparison}.

\FloatBarrier
\section{VL-LN Benchmark}
This section first defines the Interactive Instance Goal Navigation (IIGN) task in \cref{sec: task definition}. Then present VL-LN Bench, including its agent–oracle interaction design in \cref{sec: Agent–Oracle Interaction}, dataset construction pipeline in \cref{sec: dataset}, dataset statistics in \cref{sec: Statistics}, and evaluation metrics in \cref{sec: metric}.
\subsection{Task Definition}\label{sec: task definition}
The Interactive Instance Goal Navigation (IIGN) challenges a dialog-enabled embodied agent to locate a specific instance in an unknown environment. It involves two active roles: an agent and an oracle. For each episode, the agent is randomly placed in an unknown environment~\cite{szot2021habitat} and given an ambiguous instruction providing only the target category~(\eg, \textit{“Search for the chair”}).  At each step, the agent receives a visual observation $o_t$ from the environment, and chooses to either move from $a_t \in \mathcal{A}$, where
\[
\begin{aligned}
\mathcal{A}=\{&\textsc{Forward}(0.25\,\mathrm{m}),\textsc{Look-Down},~\textsc{Look-Up},\\
&~\textsc{Turn-Left}(30^\circ),~\textsc{Turn-Right}(30^\circ),~\textsc{Stop}\}
\end{aligned}
\]
or query the oracle for guidance via free-form, open-ended natural-language interactions. The oracle is assumed to have human-like prior knowledge of the environment, including (1) the detailed attributes and location of the target instance, (2) the global structure of the environment, and (3)the agent’s current state. The goal of the agent is to locate the specified instance with minimal steps under limited interactions with the oracle.

\subsection{Interaction-Supportive Design}\label{sec: Agent–Oracle Interaction}
A key design goal of IIGN is to provide an ideal task setting for evaluating the impact of dialog in embodied navigation. To this end, we construct the benchmark such that category-level instructions can be ambiguous in realistic scenes—often corresponding to \emph{multiple plausible instances}—making it difficult for an agent to reliably identify the intended target without additional information. Moreover, to enable large-scale trajectory collection and standardized evaluation, we design \emph{a scalable oracle} that can interact with the agent in natural language during both data collection and testing.

\noindent\textbf{Interaction Design.} In IIGN, the agent interacts with an oracle through free-form natural-language dialog. Although the query space is open-ended, effective information-seeking dialog in goal-oriented navigation mainly serves two purposes: \emph{identifying the target instance} and \emph{supporting efficient navigation}.

For target identification, we group queries by how they reduce uncertainty. Before grounding any candidate, the agent asks for target-level cues—intrinsic attributes (e.g., color, shape, texture, material) and spatial context (e.g., room, nearby objects)—which we call \emph{attribute} questions. These narrow the candidate set without referencing a specific observed instance. After grounding a plausible candidate, the agent asks whether it is the intended target, we call these \emph{disambiguation} questions, which verify a specific grounded instance (e.g., ``Is it the target chair?'').

To support efficient navigation, we introduce \emph{route} questions that ask how to reach the target from the agent’s current pose. We categorize them by the granularity of guidance requested: \emph{orientation-level} questions seek a coarse relative direction (the straight-line heading to the target, \eg, ``Which direction is the target?''), \emph{path-level} questions request a traversable path through the environment (\eg, ``How do I get there from here?''), and \emph{step-level} questions ask for local instructions at the current decision point (\eg, ``Should I turn left or right here?'' or ``Should I go upstairs or downstairs?''). These three levels form a natural hierarchy from global orientation to route planning to local execution.

Although the question space is open-ended, most effective information-seeking dialog in this setting concentrates on these three recurring, task-critical query types. Accordingly, we focus trajectory collection on them, as they provide the core information for IIGN.

\noindent\textbf{Oracle Support.} 
To support open-ended interaction, we implement the oracle as a GPT-4o-powered NPC. Given a query, the NPC classifies it into one of four categories: Attribute, Route, Disambiguation, and Other. It then generates a response according to the corresponding category-specific logic. Details of the NPC’s response generation are provided in the supplementary material.

\FloatBarrier
\subsection{Dataset Generation}\label{sec: dataset}
\begin{figure}[!ht]
  \centering
  \includegraphics[width=0.9\linewidth]{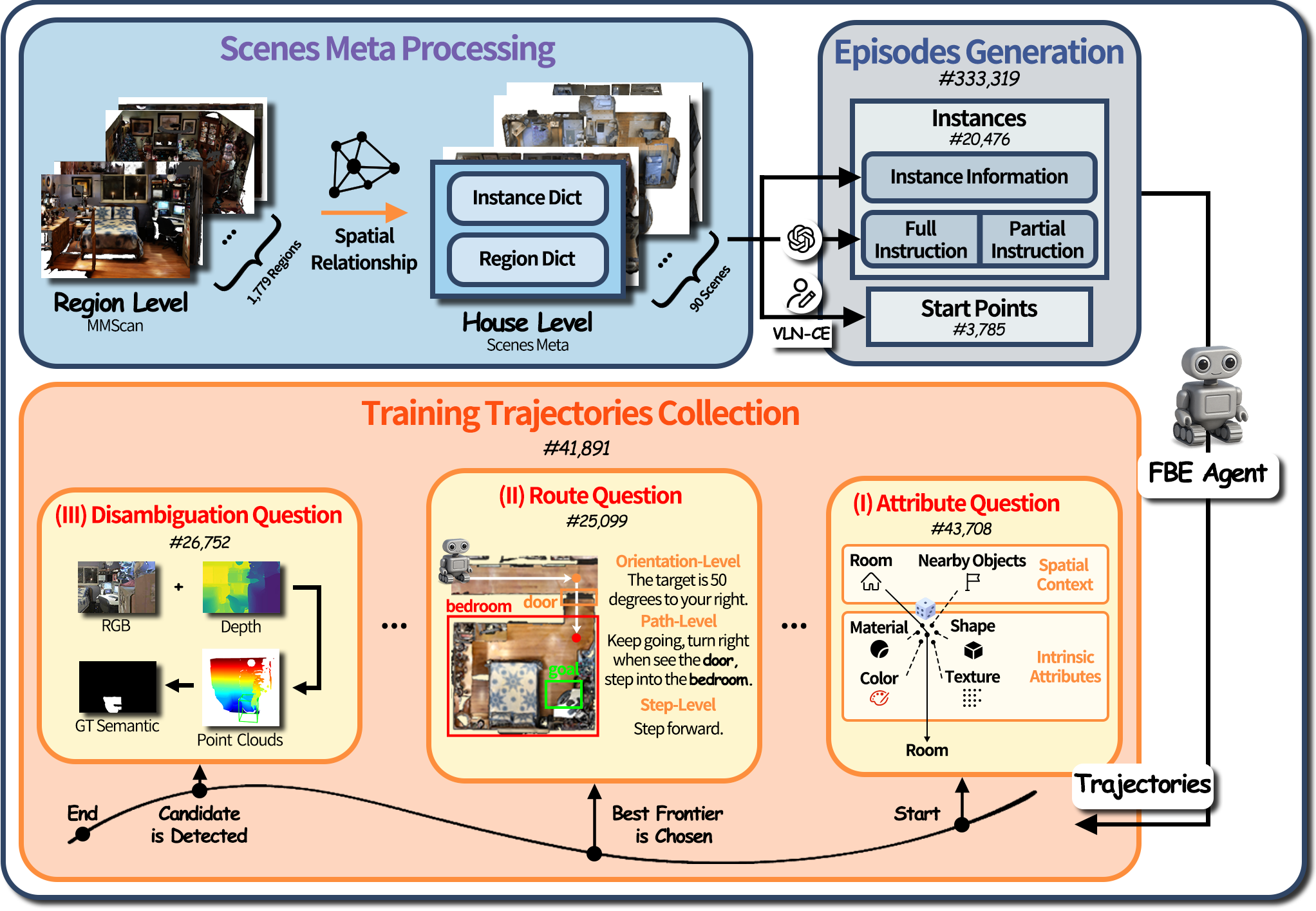}
  \caption{\textbf{Automatic pipeline for collecting dialog-augmented trajectories.} 
  The number after ``\#'' denotes the corresponding case count, and ellipses indicate that additional attribute, route, or disambiguation questions may be included.}
  \label{fig:dataprocess}
\end{figure}

\suppressfloats[t]
As shown in \cref{fig:dataprocess}, our dataset is built with a three-stage automatic pipeline. The \textsc{val} and \textsc{test} sets use only the first two stages, while the \textsc{training} set further uses the third stage to generate trajectories from episode annotations.

\FloatBarrier

\subsubsection{Scenes Metadata Processing} 
We process MP3D~\cite{Matterport3D} scenes meta using MMScan annotations~\cite{mmscan}, which provide fine-grained descriptions at both instance and region levels. Instance-level annotations include spatial properties (geometry, pose, \etc) and attributes (category, material, \etc), while region annotations specify each region’s function (\eg, dining room, study, bathroom) and its contained instances. Based on these annotations, we construct per-scene \emph{instance dictionary} and \emph{region dictionary}, unifying all region-level annotations into a single house-level index spanning the entire scene. In addition, we build a spatial-relation graph following Sr3D~\cite{Sr3d}, where nodes are instances and edges connect instances within $1\mathrm{m}$. Combined with the house-level dictionaries, this graph provides relational cues for disambiguating visually similar instances.

\subsubsection{Episodes Generation} 
Each episode is defined by three core elements: an initial agent pose, a navigation instruction, and a set of viewpoints for target instance. The \emph{initial agent poses} include starting poses from R2R-CE (72 scenes)~\cite{vln-ce} and our manually annotated plausible positions in 18 scenes. For every instance, we provide two \emph{instruction} variants: a partial, category-only instruction mentioned above (\eg, \textit{`Search for the chair.''}) and a full description that uniquely identifies the target among all instances in the scene (\eg, \textit{Locate the deep grey chair with black backrest, standing upright on a wooden floor near a computer and a tv in the bedroom.''}). The full instruction supports a non-interactive instance goal navigation setting (\ie IGN), while the partial instruction is used in IIGN. We generate a set of \emph{viewpoints} for each instance, which serve as expected `Stop'' locations. The episode is considered to be successful if the agent stops within 0.25 $ m$ of any viewpoint of the instance. Combining an initial agent pose with the instance’s instruction and its viewpoints defines an episode. More details on generating the initial agent poses, instructions, and viewpoints are provided in the supplementary material.

\subsubsection{Training Trajectories Collection} We use an agent in Habitat~\cite{habitat19iccv,szot2021habitat}, equipped with an RGB--D camera and odometry, to generate trajectory data from annotated episodes. The agent follows a frontier-based exploration (FBE) policy and is equipped with a ground-truth object detector. Specifically, at each step, the agent selects either the frontier closest to the previously visited one or the frontier with the smallest geodesic distance to the target instance, balancing local exploration and goal-directed guidance. Detailed implementation of FBE is provided in the supplementary material. Importantly, unlike in VLN, shortest-path supervision is often inadequate for goal-oriented navigation, because the instruction specifies what to find rather than how to get there, making active exploration necessary. Consequently, training on exploration-aware trajectories is crucial. As shown in \cref{sec:ablation}, models trained on such data consistently outperform those trained only on shortest-path trajectories.

To collect dialog of the three types mentioned above, we define question triggers during navigation. An \emph{Attribute} question is asked at the start of each trajectory, with the queried attribute randomly sampled from the set in \cref{fig:dataprocess}. A \emph{Route} question is issued when the geodesically closest frontier (best frontier) is selected. A \emph{Disambiguation} question is triggered when the agent detects an instance of the target category that matches the known attribute. Here, an instance is considered ``detected'' if its ground-truth semantic label appears near the image center and it lies within $3\mathrm{m}$ of the agent. For each question type, we use multiple semantically consistent templates to increase dialog diversity. Finally, we log all sensor observations and the full dialog history to construct dialog-augmented trajectories.

\subsubsection{Trajectories Filtering}
We filter trajectories in three steps to improve annotation reliability and reduce training noise.
\emph{First}, we discard any trajectory that contains collision events, ensuring the policy is trained on collision-free behaviors.
\emph{Second}, we remove trajectories with more than 13 consecutive turning actions, since each turn is $30^\circ$, reorientation requires at most 12 turns thus longer sequences are redundant.
\emph{Finally}, we enforce target-centered terminal observations by extracting the final RGB frames, 
and keep a trajectory only if the target appears within the central region of the terminal image, defined as the inner $3/4 \times 3/4$ area after cropping a $1/8$ margin from each border.


\subsection{Data Statistics}\label{sec: Statistics}
Our dataset covers 112 object categories and contains 20,476 annotated instances from 90 MP3D scenes. For each scene, we sample a set of navigable start poses, with an average of 42 per scene. Pairing these start poses with annotated target instances yields 333,319 episodes in total. Following the VLN-CE split protocol~\cite{vln-ce}, we partition the scenes into 61/15/14 for training, validation, and testing, respectively, resulting in 246,433 training episodes, 86,386 validation episodes, and 500 test episodes. Based on the training episodes, we further collect 41,891 IIGN trajectories covering all training scenes and object categories, where each trajectory couples navigation with question-answer interaction. Detailed trajectory statistics are provided in the supplementary material. In addition, for our experiments, we collect 17,224 IGN trajectories and 20,833 ObjectNav trajectories using our data generation pipeline and we further filtered the IIGN trajectories to 11,661 with balanced scenes and target categories to mitigate imbalance.

\subsection{Evaluation Metric}\label{sec: metric}
In addition to the standard navigation related metrics—Success Rate (SR), Success Rate weighted by Path Length (SPL), Oracle Success Rate (OS), and Navigation Error (NE), we introduce the Mean Success Progress (MSP) metric to specifically evaluate dialog utility. Given a maximum dialog allowance of $n$ turns, we first compute the baseline success rate without dialog, denoted as $s_0$. Then measure the success rate under increasing dialog budgets, yielding $s_1, ..., s_n$. For each budget, we calculate the success improvement relative to the baseline, \ie, 
$(s_i - s_0), 0 <i\leq n$. The MSP score is defined as the mean of these improvements across all dialog budgets, \ie, \(\mathrm{MSP} = \frac{1}{n}\sum_{i=1}^{n}(s_i - s_0)\). This metric captures both the effectiveness and efficiency of dialog utility. It measures the average improvement in navigation success brought by dialog, while favoring larger gains achieved with fewer dialog turns. In this paper, we set $n=5$.


\FloatBarrier
\section{Experiments}
In this section, we first describe our experimental setup in \cref{sec:exp_set}. We then analyze why instance-goal navigation is intrinsically challenging and how dialog improves performance in \cref{sec:main_results}. Finally, we validate the soundness of our benchmark and highlight some interesting findings through ablation studies in \cref{sec:ablation} and real-world experiments in \cref{sec:real_world}.
\subsection{Experimental Setup}\label{sec:exp_set}
\noindent\textbf{Evaluated task.}
In addition to IIGN, we benchmark Instance Goal Navigation (IGN) without dialog, as most prior methods lack dialog capabilities. To enable the agent to identify the target instance, we provide a \emph{full instruction} in IGN, i.e., a complete and unambiguous description of the target.

\noindent\textbf{Baselines.}
We evaluate four baselines on the two VL-LN Bench tasks, IGN and IIGN. The baselines are: (i) a greedy frontier-based exploration (FBE) agent with an open-vocabulary detector based on Grounded SAM~2~\cite{GroundingSam2}, (ii) VLFM~\cite{vlfm}, (iii) AIUTA~\cite{taioli2024collaborative} an interactive instance goal navigation method that equipped with an LLM--VLM Self-Questioner, and (iv) our VLLN model, which is initialized from \textit{Qwen3-VL-4B-Instruct}~\cite{qwen3technicalreport} and trained with the InternVLA-N1~\cite{internvla-n1} recipe on ObjectNav (20{,}833), IGN (17{,}224), and IIGN (11{,}661) trajectories collected by our pipeline. We report VLLN in both dialog-enabled and non-dialog settings. We also train three ablated variants, VLLN-O, VLLN-I, and VLLN-II, using only ObjectNav, IGN, or IIGN data, respectively. Additional implementation details and real-world experimental settings are provided in the supplementary material.

\FloatBarrier
\subsection{Main Results}\label{sec:main_results}
\suppressfloats[t]

\begin{table}[!ht]
\centering
\footnotesize
\setlength{\tabcolsep}{4pt}
\renewcommand{\arraystretch}{1.15}

\caption{\textbf{Results of the VL-LN Benchmark.}}
\label{tab:benchmark}

\resizebox{\linewidth}{!}{
\begin{tabular}{@{} c c | c | c c c c c | c c c c c @{}}
\hline
\multirow{2}{*}{Task} & \multirow{2}{*}{Method}
& \multicolumn{1}{c|}{\multirow{2}{*}{w/ Dialog}}
& \multicolumn{1}{c}{\multirow{2}{*}{SR\,$\uparrow$}}
& \multicolumn{1}{c}{\multirow{2}{*}{SPL\,$\uparrow$}}
& \multicolumn{1}{c}{\multirow{2}{*}{OS\,$\uparrow$}}
& \multicolumn{1}{c}{\multirow{2}{*}{NE\,$\downarrow$}}
& \multicolumn{1}{c}{\multirow{2}{*}{MSP\,$\uparrow$}}
& \multicolumn{3}{|c}{Detection-related}
& \multirow{2}{*}{Expl.}
& \multirow{2}{*}{ST} \\
\cline{9-11}
& & & & & & & & MD & WD & AD & & \\
\hline
\multirow{6}{*}{IIGN}
& FBE      & \xmark & 9.2  & 5.09  & 12.0 & 11.51 & - & 1.4\% & 37.8\% & 31.0\% & 19.0\% & 1.6\% \\
& VLFM\cite{vlfm}    & \xmark & 8.8 & 5.63  & 13.6 & 11.11 & - & 7.6\% & 21.4\% & 33.2\% & 26.4\% & 2.6\% \\
& AIUTA\cite{taioli2024collaborative}   & \cmark  & 4.0 & 2.07 & 27.6 & 10.53 & 1.04 & 25.2\% & 18.8\% & 9.2\% & 39.6\% & 3.2\% \\
& VLLN  & \xmark & 19.6 & 12.32 & 28.6 & 9.70 & - & 5.8\% & 14.8\% & 36.8\% & 16.8\% & 6.2\% \\
& VLLN   & \cmark & \bfseries 25.4 & \bfseries 16.18 & \bfseries 35.2 & \bfseries 8.61 & 4.12 & 5.8\% & 13.0\% & 33.6\% & 15.6\% & 6.6\% \\
\hline
\multirow{6}{*}{IGN}
& FBE & \xmark & 5.8  & 3.45  & 13.8 & 12.18 & - & 7.4\% & 36.2\% & 21.6\% & 26.6\% & 2.4\% \\
& VLFM\cite{vlfm} & \xmark & 9.6 & 6.13  & 15.4 & 11.19 & - & 9.2\% & 22.6\% & 32.6\% & 23.8\% & 2.2\% \\
& AIUTA\cite{taioli2024collaborative}   & \cmark  & 5.6 & 3.24 & 28.4 & 10.62 & 0.86 & 23.2\% & 21.2\% & 8.8\% & 38.8\% & 2.4\% \\
& VLLN  & \xmark & 27.0 & 17.02 & 36.2 & 9.10 & - & 4.6\% & 14.6\% & 32.8\% & 13.0\% & 8.0\% \\
& VLLN  & \cmark & \bfseries 30.2 & \bfseries 18.73 & \bfseries 40.6 & \bfseries 7.72 & 3.24 & 5.0\% & 12.2\% & 30.8\% & 14.0\% & 7.8\% \\
\hline
\end{tabular}
}
\begin{tablenotes}[flushleft]
\scriptsize
\item \textbf{w/ Dialog}: the agent can interact with an NPC by asking questions during navigation. \textbf{MD} = Missed Detection; \textbf{WD} = Wrong Detection; \textbf{AD} = Ambiguous Detection; \textbf{Expl.}\ = Exploration Fail; \textbf{ST} = Stop Fail.
\end{tablenotes}
\end{table}

\noindent\textbf{Key Challenges for IGN and IIGN.}
We evaluate a range of agents, including both zero-shot and trained methods, on the IGN and IIGN tasks in our benchmark, and observe a substantial performance drop when moving from ObjectNav~\cite{objectnav} to these instance-level settings. As shown in ~\cref{tab:benchmark}, prior method VLFM~\cite{vlfm} achieve 36.4\% success rate (SR) on ObjectNav~\cite{objectnav}, but degrade markedly on IGN and IIGN. A similar trend is also observed in our model: in ~\cref{tab:role ablation}, VLLN drops from 69.2\% SR on ObjectNav to 30.2\% on IGN and 25.4\% on IIGN, despite co-training on collected ObjectNav, IGN, and IIGN trajectories.

\begin{figure}[!ht]
    \centering
    \includegraphics[width=0.55\linewidth]{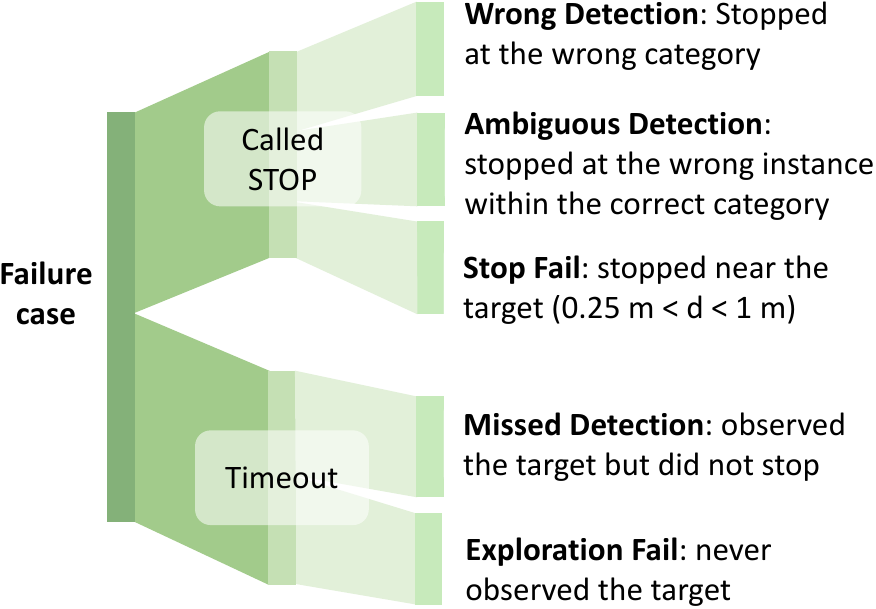}
    \caption{Failure taxonomy used in our error analysis.}
    \label{fig:failure_analysis}
\end{figure}
\FloatBarrier

We attribute this gap to two core challenges shared by IGN and IIGN.
First, \emph{long-horizon exploration} becomes more demanding. ObjectNav\cite{objectnav} succeeds upon reaching any instance of the target category (\eg, any of the seven ``chair'' candidates in ~\cref{fig:teaser}), whereas IGN and IIGN require finding a specific instance, which often entails broader and longer exploration.
Second, \emph{attribute--image alignment} is difficult: agents must map fine-grained textual attributes to the correct visual instance, but they often miss or misread attributes and stop at same-category distractors instead of the target.

To better understand model failures, we partition the failure cases in \cref{fig:failure_analysis} into five categories and analyze their distributions. The results indicate that \emph{attribute--image alignment} is the primary bottleneck in both IGN and IIGN. As shown in~\cref{tab:role ablation}, there is a clear gap between ObjectNav~\cite{objectnav} and instance-goal navigation, and this gap is driven predominantly by \emph{Ambiguous Detection}, which is the largest error component (27.8\%--34.6\% on IGN/IIGN), while the remaining failure types vary more modestly. This indicates that even with detailed target descriptions, the model often fails to correctly ground fine-grained attributes to the intended instance when multiple same-category candidates are present. 

Beyond the shared challenges above, IIGN introduces an additional challenge of \emph{effective questioning}. As shown in \cref{tab:benchmark}, the SR drop of VLLN w/ Dialog from IGN to IIGN mainly stems from detection-related errors, indicating that dialog information remains less effective than full instructions for identifying the target among multiple candidates. Therefore, IIGN requires agents to use dialog effectively to obtain the most informative disambiguating cues.


\begin{figure}[!ht]
\centering
\includegraphics[width=\linewidth]{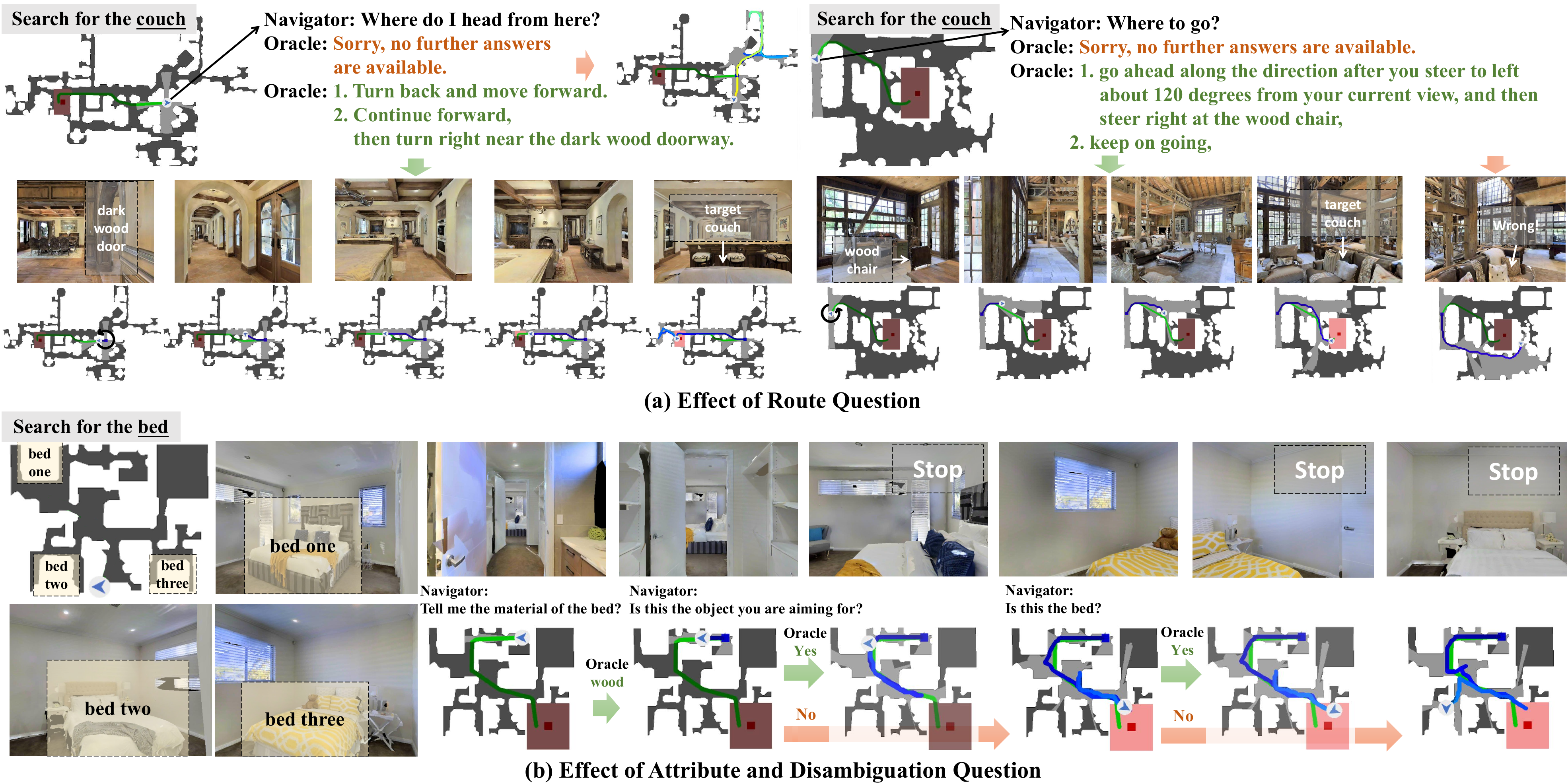}
\caption{\textbf{Qualitative examples of how dialog improves navigation.}
(a) \textbf{Effect of route question.} After receiving route guidance, the agent follows the instructed turns, steadily approaches the target, and ultimately succeeds; without answers, the agent explores blindly and eventually reaches the time limit.
(b) \textbf{Effect of attribute and disambiguation question.} When multiple candidates from the same category are present, attribute questions provide discriminative cues, while disambiguation questions steer the agent away from incorrect instances that have not yet been ruled out.}

\label{fig:dialog_help}
\end{figure}
\FloatBarrier
\noindent\textbf{Dialog Helps.} To examine the impact of dialog, we compare dialog-enabled agent with non-dialog baselines on both IGN and IIGN (\cref{tab:benchmark}), and further visualize representative episodes to illustrate how dialog alters navigation behaviors (\cref{fig:dialog_help}). As shown in \cref{tab:benchmark}, VLLN w/ Dialog achieves the best overall performance on both tasks. Compared with its non-dialog counterpart, it improves SR and SPL by 3.2--5.8\% and 1.71--3.86\%, respectively, indicating that actively seeking additional information yields consistent gains.

These improvements can be understood through two factors: dialog enhances navigation efficiency by providing actionable guidance that reduces unnecessary exploration and encourages more directed trajectories; and it strengthens instance-level disambiguation by supplying discriminative cues that help the agent rule out incorrect candidates and re-route when encountering misleading detections. Consistent with the efficiency explanation, \cref{fig:dialog_help}(a) shows that when the agent asks a route question and receives guidance, it follows the instructed actions and steadily moves closer to the target, whereas without answers it explores blindly and ultimately fails. Meanwhile, \cref{fig:dialog_help}(b) illustrates that answers to disambiguation questions can directly determine the agent's stopping position. In this example, different answers lead the agent to stop beside different beds.  The benefits are also reflected quantitatively in the failure analysis: on both tasks, VLLN reduces WD and AD compared with non-dialog VLLN (IGN: WD 12.2\% vs.\ 14.6\%, AD 30.8\% vs.\ 32.8\%; IIGN: WD 13.0\% vs.\ 14.8\%, AD 33.6\% vs.\ 36.8\%). Moreover, its higher MSP on IIGN than on IGN (4.12 vs.\ 3.24) suggests that dialog is particularly helpful under the ambiguous setting.

\FloatBarrier
\subsection{Ablation Studies}\label{sec:ablation}
\suppressfloats[t]

\noindent\textbf{Data Collection Strategy.} We compare two trajectory collection strategies (shortest path vs. frontier exploration in our pipeline). Overall, models fine-tuned on exploration trajectories consistently outperform those trained on shortest-path trajectories. \Cref{table:Data Collection Strategy} summarizes the results and failure statistics. On ObjectNav, frontier-based exploration improves performance to 72.4\% SR and 81.1\% OS, compared to 60.5\% SR and 69.3\% OS with shortest-path data. On IGN, frontier-based exploration also yields higher SR and OS, with gains of 6.4\% SR and 6.8\% OS over shortest path trajectories.

Our results suggest that exploratory trajectories are more effective for supervised fine-tuning than shortest-path trajectories. Because shortest-path trajectories are typically shorter, the model may learn to associate stopping with a limited number of steps rather than with actually reaching the target. We observe that, in both ObjectNav and IGN, models trained with frontier-exploration trajectories exhibit higher exploration error (Expl.) but substantially lower wrong detection (WD) than those trained with shortest-path data. This suggests that, when the target has not been found, frontier-exploration training encourages the agent to continue searching, whereas shortest-path training makes the agent more prone to stopping prematurely before getting close to the correct target.

\begin{table}[!ht]
\centering
\caption{ObjectNav and IGN results under different data collection strategies.}
\label{table:Data Collection Strategy}
\footnotesize
\setlength{\tabcolsep}{4pt}
\renewcommand{\arraystretch}{1.15}
\resizebox{\linewidth}{!}{
\begin{tabular}{@{} c c | c c c c | c c c c c @{}}
\hline
\multirow{2}{*}{Task} & \multirow{2}{*}{Method}
& \multicolumn{1}{c}{\multirow{2}{*}{SR\,$\uparrow$}}
& \multicolumn{1}{c}{\multirow{2}{*}{SPL\,$\uparrow$}}
& \multicolumn{1}{c}{\multirow{2}{*}{OS\,$\uparrow$}}
& \multicolumn{1}{c|}{\multirow{2}{*}{NE\,$\downarrow$}}
& \multicolumn{3}{c}{Detection-related}
& \multirow{2}{*}{Expl.}
& \multirow{2}{*}{ST} \\
\cline{7-9}
& & & & & & MD & WD & AD & & \\
\hline
\multirow{2}{*}{Objectnav}
& Shortest Path      & 60.5  & 27.23  & 69.3 & 2.54 & 2.4\% & 28.4\% & - & 4.7\% & 4.0\% \\
& Frontier Exploration     & 72.4 & 30.36  & 81.1 & 1.57 & 5.7\% & 8.6\% & - & 10.7\% & 2.6\% \\

\hline
\multirow{2}{*}{IGN}
& Shortest Path      & 26.0  & 15.40  & 32.6 & 9.68     & 2.0\% & 29.0\% & 30.8\% & 7.0\% & 5.2\% \\
& Frontier Exploration     & 32.4 & 19.76  & 39.4 & 7.96 & 4.6\% & 12.6\% & 27.8\% & 16.2\% & 6.4\% \\
\hline
\end{tabular}}
\begin{tablenotes}[para]
\scriptsize
\end{tablenotes}
\end{table}

\noindent\textbf{Co-training vs. Task-specific.}
To study whether a single policy can generalize across all three goal-oriented navigation tasks, we compare task-specific training models (VLLN-O, VLLN-I, VLLN-II) with the co-training model VLLN on ObjectNav, IGN, and IIGN. As shown in~\cref{tab:role ablation}, co-training induces only minor performance changes across tasks, suggesting limited negative transfer. In particular, ObjectNav and IGN exhibit modest variations, while IIGN remains essentially unchanged, indicating that joint training across the three tasks is feasible. These results support the prospect of a unified goal-oriented navigation agent that can handle both category-level and instance-level targets under either unambiguous or ambiguous settings.

Beyond task-level stability, we observe cross-task transfer of interaction ability. Although the IGN training trajectories contain only navigation actions, co-training with IIGN enables the model to acquire question-asking behavior that also manifests under the IGN setting. Under our evaluation protocol, this transferred interaction capability improves IGN performance, increasing SR from 27.0\% to 30.2\%. This indicates that interaction behaviors learned from interactive trajectories can transfer to tasks trained without dialog supervision and still improve navigation success.

Notably, all trajectories for the three tasks are collected using our automatic pipeline, demonstrating that the pipeline supports both navigation and dialog supervision and can serve as a unified data engine for goal-orient navigation.

\begin{table}[tb]
\centering
\caption{Results on ObjectNav, IGN, and IIGN under task-specific training and multi-task co-training.}
\label{tab:role ablation}
\footnotesize
\setlength{\tabcolsep}{4pt}
\renewcommand{\arraystretch}{1.15}
\resizebox{\linewidth}{!}{
\begin{tabular}{@{} c c | c | c c c c | c c c c c @{}}
\hline
\multirow{2}{*}{Task} & \multirow{2}{*}{Method} & \multirow{2}{*}{\makecell{w/\\Dialog}}
& \multicolumn{1}{c}{\multirow{2}{*}{SR\,$\uparrow$}}
& \multicolumn{1}{c}{\multirow{2}{*}{SPL\,$\uparrow$}}
& \multicolumn{1}{c}{\multirow{2}{*}{OS\,$\uparrow$}}
& \multicolumn{1}{c|}{\multirow{2}{*}{NE\,$\downarrow$}}
& \multicolumn{3}{c}{Detection}
& \multirow{2}{*}{Expl.}
& \multirow{2}{*}{ST} \\
\cline{8-10}
& & & & & & & MD & WD & AD & & \\
\hline
\multirow{2}{*}{Objectnav}
& VLLN-O  & \xmark  & 72.4 & 30.36 & 81.1 & 1.57 & 5.7\% & 8.6\%  & -      & 10.7\% & 2.6\% \\
& VLLN   & \xmark  & 69.2 & 27.61 & 77.5 & 1.83 & 4.7\% & 15.2\% & -      & 7.9\%  & 3.0\% \\
\hline
\multirow{3}{*}{IGN}
& VLLN-I  & \xmark  & 32.4 & 19.76 & 39.4 & 7.96 & 4.6\% & 12.6\% & 27.8\% & 16.2\% & 6.4\% \\
& VLLN   & \xmark  & 27.0 & 17.02 & 36.2 & 9.10 & 4.6\% & 14.6\% & 32.8\% & 13.0\% & 8.0\% \\
& VLLN    & \cmark & 30.2 & 18.73 & 40.6 & 7.72 & 5.0\% & 12.2\% & 30.8\% & 14.0\% & 7.8\% \\
\hline
\multirow{2}{*}{IIGN}
& VLLN-II  & \cmark & 25.0 & 15.63 & 33.8 & 9.04 & 4.8\% & 18.2\% & 34.6\% & 11.4\% & 6.0\% \\
& VLLN    & \cmark & 25.4 & 16.18 & 35.2 & 8.61 & 5.8\% & 13.0\% & 33.6\% & 15.6\% & 6.6\% \\
\hline
\end{tabular}}
\begin{tablenotes}[para]
\scriptsize
\end{tablenotes}
\end{table}

\begin{table}[tb]
\centering
\caption{Results of cross-role ablation under different navigator--oracle combinations in IIGN.}
\label{tab:cross_role_eval}
\footnotesize
\setlength{\tabcolsep}{4pt}
\renewcommand{\arraystretch}{1.15}

\begin{tabular}{l
                S[table-format=2.1]
                S[table-format=2.2]
                S[table-format=2.1]
                S[table-format=2.2]
                S[table-format=2.2]}
\toprule
\multicolumn{1}{l}{Navigator--NPC} &
\multicolumn{1}{c}{SR\,$\uparrow$} &
\multicolumn{1}{c}{SPL\,$\uparrow$} &
\multicolumn{1}{c}{OS\,$\uparrow$} &
\multicolumn{1}{c}{NE\,$\downarrow$} &
\multicolumn{1}{c}{Avg.\ turns} \\
\midrule
Human--Human & 93.0 & 57.30 & 95.0 & 0.31 & 2.04 \\
Human--GPT   & 91.0 & 49.88 & 94.0 & 0.69 & 9.72 \\
VLLN--Human  & 25.0 & 15.63 & 33.4 & 7.52 & 1.86 \\
VLLN--GPT    & 25.4 & 16.18 & 35.2 & 8.61 & 1.98 \\
\bottomrule
\end{tabular}
\end{table}

\noindent\textbf{Navigator--NPC Role.} In the IIGN task, performance depends on two key roles: the \emph{navigator} and the \emph{NPC}. To validate the soundness of our benchmark and the reliability of our evaluation protocol, we conduct role ablations by replacing each role with a human. Our default setting uses the VLLN model as the navigator and the GPT-based NPC for automatic evaluation (denoted as \emph{VLLN-GPT}). Detailed results are reported in \cref{tab:cross_role_eval}.

Our results suggest that the GPT-based NPC is a reliable proxy for human performance in IIGN. Replacing it with a human NPC (\emph{VLLN-Human}) yields performance comparable to \emph{VLLN-GPT}, indicating similar quality to human answers. When the navigator is replaced by a human while retaining the GPT-based NPC (\emph{Human-GPT}), the performance also remains close to human-human evaluation. Together, these results show that the GPT-based NPC is not only a reliable substitute for human NPCs in automatic evaluation, but also generalizes well to free-form, open-ended dialog from human navigators.

There remains a marked performance gap between our model and human navigators. Human navigators achieve over 90\% SR, which is far higher than the 25\%--25.4\% SR of VLLN. To ensure the reliability of the test set, we verified that all episodes are valid, and further analyzed the failure cases in the \emph{Human-Human} and \emph{Human-GPT} settings to summarize the reasons for human errors (see the supplementary material).

\FloatBarrier
\subsection{Real-world Experiments}\label{sec:real_world}
\begin{figure}[!ht]
\centering
\includegraphics[width=\linewidth]{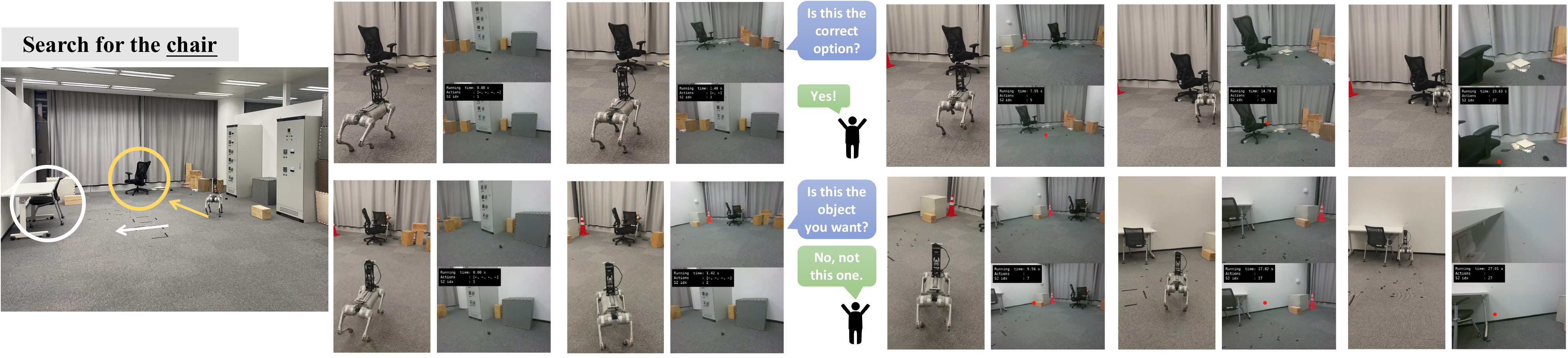}
\caption{\textbf{ Real-world deployment.} A Unitree Go2 robot searches for a specific chair under different human responses. At each step, we show three views: a third-person view, the robot view, and the decision frame.}
\label{fig:failure_case}
\end{figure}

\suppressfloats[t]
In this experiment, we place two chairs in a room and provide different human responses to guide the robot to different target chairs. \cref{fig:failure_case} shows step-wise snapshots of the navigation and interaction process; the full rollout is provided in the supplementary video.

\FloatBarrier

\noindent\textbf{Qualitative Analysis.}
These results demonstrate the feasibility of interactive navigation in real-world scenarios. When the robot observes a chair that matches the instruction but is not sure whether it is the target, it proactively asks a disambiguation question. Different human responses lead the robot to different target chairs, demonstrating that the robot can actively seek clarification when uncertain and use human feedback to complete the task.

\FloatBarrier
\section{Conclusion}
This paper investigates the Interactive Instance Goal Navigation (IIGN) task and introduces the VL-LN bench, which includes an automatic data collection pipeline and a long-horizon dataset comprising 41k collected dialogue-augmented trajectories for training, along with an evaluation protocol involving a reliable oracle for agent assessment. Our experiments highlight the challenges of IGN and IIGN and show that active dialog consistently improves performance on both tasks. Ablations confirm the value of our collected data and demonstrate that the proposed oracle provides scalable assistance comparable to human support, thereby establishing VL-LN Bench as a practical testbed for dialog-enabled embodied navigation.
\FloatBarrier

\bibliographystyle{plainnat}
\bibliography{main}

\clearpage
\beginappendix
\renewcommand{\theHsection}{appendix.\Alph{section}}
\renewcommand{\theHsubsection}{\theHsection.\arabic{subsection}}
\section{More Implementation Details}
\subsection{NPC Implementation} \label{sec: npc}
\begin{figure}[!ht]
\centering
\includegraphics[clip,width=0.9\linewidth]{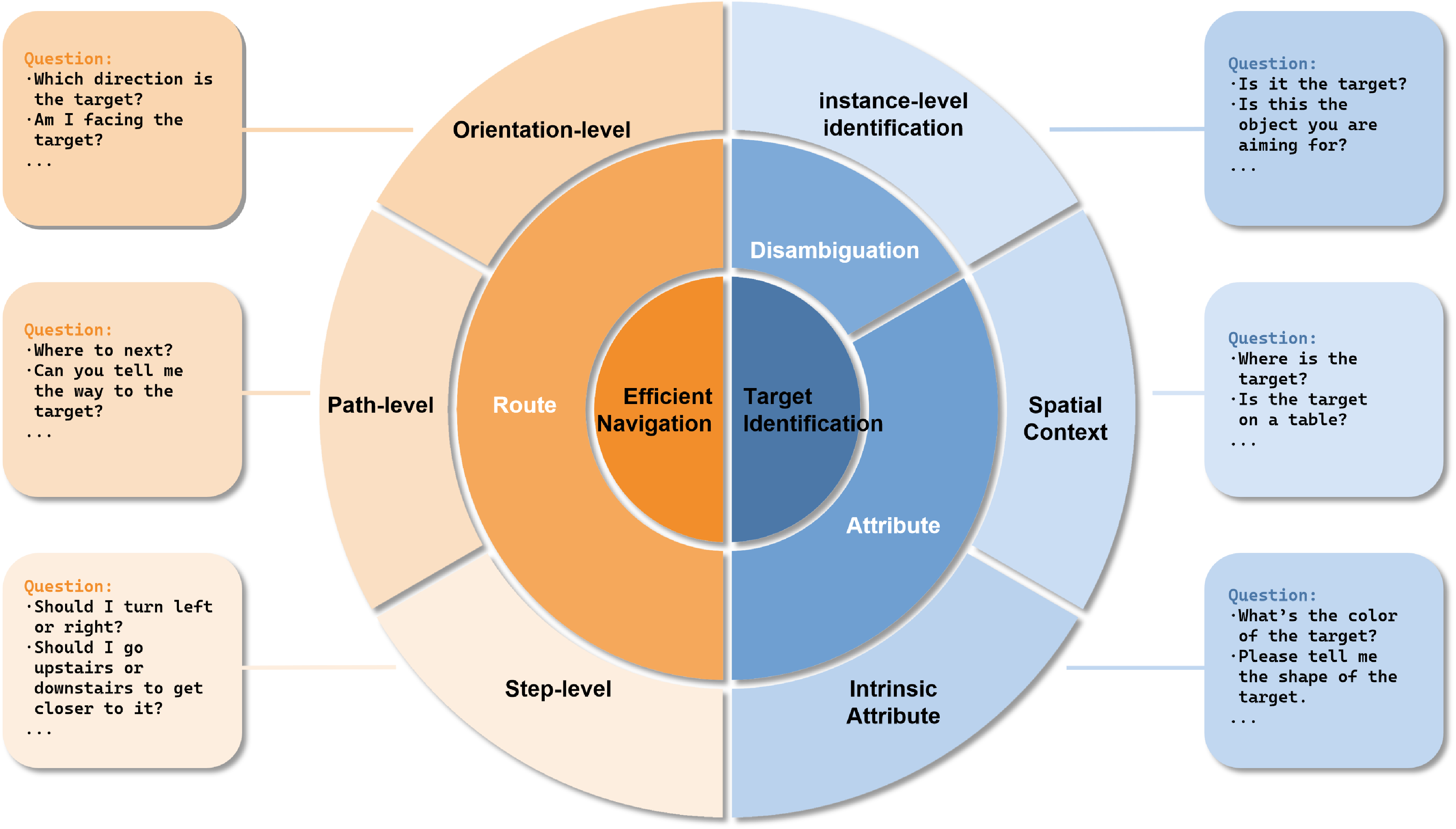}
    \caption{\textbf{Hierarchical taxonomy of question types in VL--LN Bench.} The inner ring summarizes the two main purposes of dialog: t\emph{arget identification} and \emph{efficient navigation}. The second ring shows the corresponding question types, including \emph{attribute}, \emph{disambiguation}, and \emph{route}. The outer ring further breaks them down into finer-grained subtypes: \emph{intrinsic attribute} and \emph{spatial context} for attribute questions, \emph{instance-level identification} for disambiguation questions, and \emph{orientation-level}, \emph{path-level}, and \emph{step-level} for route questions. The surrounding text boxes provide representative examples of each subtype.}
    \label{fig:comprehensive}
\end{figure}

Although the question space in interactive instance-goal navigation is open-ended, we propose a comprehensive taxonomy of dialog types that covers most practical information-seeking queries in this setting, see~\cref{fig:comprehensive}. The categories are designed to be functionally distinct and largely orthogonal to one another, including target identification and navigation support, with finer-graineds subtypes such as attribute, disambiguation, and route questions.

Building upon this taxonomy, we develop an NPC powered by GPT-4o \cite{gpt4o}, related prompt can be found in~\cref{tab: npc_prompt}. Given a query, it first classifies the query into one of four types—\emph{Attribute}, \emph{Route}, \emph{Disambiguation}, or \emph{Other}. It then produces a response using type-specific logic:
\begin{itemize}[leftmargin=*, itemsep=2pt, topsep=2pt]
    \item \textbf{Attribute.} 
    The NPC answers questions about the target attributes using instance-level metadata of the target instance, and returns a natural-language response generated by GPT-4o~\cite{gpt4o}.
    
    \item \textbf{Route.}
    The NPC provides three granularities of route guidance according to the level of route question. 
    For \emph{orientation-level} queries, it computes the left or right turning angle from the agent’s current heading to the target direction. 
    For \emph{path-level} queries, it converts the shortest path from the agent’s current pose to the target into natural-language instructions: it keeps only the first \(4\,\mathrm{m}\) of the path, abstracts it into waypoints at major turns or room transitions, anchors each waypoint to nearby salient objects, and verbalizes the resulting waypoint sequence as guidance. 
    For \emph{step-level} queries, it returns the next action suggested by the \texttt{ShortestPathFollower} implemented in HabitatSim~\cite{habitat19iccv}.
    
    \item \textbf{Disambiguation.} 
    The NPC answers whether the currently observed candidate is the target. 
    It returns ``yes'' if the target is centered in the current view and within \(3\,\mathrm{m}\) of the agent, and ``no'' otherwise.
    
    \item \textbf{Other.} 
    For queries outside the three core types, the NPC attempts to answer whenever the request can be grounded in its privileged information, including target metadata, task progress, and the agent’s current state. If the required information is unavailable, it returns the default response: ``Sorry, I cannot answer your question now.''
\end{itemize}

\begin{table}[htbp!]
\caption{\textbf{NPC prompts.} Prompt~1 is used for question-type identification, whereas Prompts~2--4 are used to generate responses for \emph{Attribute}, \emph{Route}, and \emph{Other} questions, respectively. Disambiguation questions are answered deterministically with ``Yes'' or ``No''; therefore, GPT-4o is invoked only once (for type identification) in this case. The placeholders \{goal\_information\}, \{heading\}, \{path\_description\}, \{next\_step\} and \{target\_detected\} are obtained by the logic described in \cref{sec: npc}}
\label{tab: npc_prompt}
\vspace{-2mm}
\small
\setlength{\tabcolsep}{6pt}
\renewcommand{\arraystretch}{1}

\begin{tabularx}{\linewidth}{@{}p{0.14\linewidth}Y@{}}
\toprule
\textbf{Prompt 1}
& You are a helpful assistant in classifying navigation task. You will be given a question among the following four types:\\
& 1. Disambiguation: This question is asked to check whether the agent has found the goal object. Like ``Is it the object you are looking for?''\\
& 2. Route: This question is asked to get how to reach the target from the agent’s current pose. Like ``Which direction is the target?'', ``Where should I go now?'', ``Should I turn left or right here?''\\
& 3. Attribute: This question is asked to get more information about the goal object. Like ``Where is the goal object?'', ``What is the color of the goal object?''\\
& 4. Other: If the question does not fall into any of the three types above.\\ \\

& You need to classify the question into one of the four types. Only output the name of the type(disambiguation, route, attribute, other). Don't be verbose.\\ \\

& Here is the question you need to classify\\
& QUESTION: \{question\}
\\
\midrule

\textbf{Prompt 2}
& You are a helpful assistant in answering the question. Here follows the ground truth information you know. You need to answer the question based on the ground truth information. \\ \\
& Here is the ground truth information about the goal object \\
& Goal Information: \{goal\_information\} \\ \\
& Here is the question you need to answer \\
& QUESTION: \{question\}
\\
\midrule

\textbf{Prompt 3}
& You are a helpful assistant in answering the question. Here follows the ground truth information you know. You need to answer the question based on the ground truth information. \\ \\
& Here is the ground truth information about next move \\
& Heading to Target: \{heading\} \\
& Correct Path: \{path\_description\} \\
& Next Step: \{next\_step\} \\ \\
& Here is the question you need to answer \\
& QUESTION: \{question\}
\\
\midrule

\textbf{Prompt 4}
& You are a helpful assistant in answering the question. Here follows the ground truth information you know. You need to answer the question based on the ground truth information. \\ \\

& Here is the ground truth information about the goal object, next move and task state\\
& Goal Information: \{goal\_information\} \\
& Heading to Target: \{heading\} \\
& Correct Path: \{path\_description\} \\
& Next Step: \{next\_step\} \\
& Target Detected: \{target\_detected\} \\ \\
& Here is the question you need to answer \\
& QUESTION: \{question\}
\\
\bottomrule
\end{tabularx}
\end{table}

\FloatBarrier
\subsection{Episode Generation}
\suppressfloats[t]
Each episode is defined by three components: the initial agent pose, the instruction, and the target viewpoints.

We manually annotated initial agent poses in 18 scenes. For these scenes, we randomly sampled navigable points and snapped the agent to each candidate location, accepting it as a valid initial pose only if it lay on the navmesh, was collision-free, and remained within the scene bounds. Combined with the original initial poses provided in VLN-CE~\cite{vln-ce}, this process yielded 3,785 initial agent poses in total.

The instructions include partial instructions (for IIGN) and full instructions (for IGN). Partial instructions are straightforward to generate because they require only the target category. In contrast, full instructions are more challenging, as they must identify a specific target instance within the entire scene. To generate full instructions, we leverage house-level dictionaries and a spatial-relation graph to select discriminative instance attributes and relations, and then prompt GPT-4o~\cite{gpt4o} to produce natural-language instructions (see \cref{tab:instruction_prompt}).

To obtain instance viewpoints, we first expand each instance's 3D bounding box by $0.6\,\mathrm{m}$ in all directions, based on an empirically chosen reasonable viewing distance. We then treat all navigable points within the expanded region as candidate viewpoints for that instance.

\begin{table}[htbp!]
\caption{\textbf{Prompt for full instruction generation.} The placeholder values are retrieved from the instance dictionary.}
\label{tab:instruction_prompt}
\vspace{-2mm}
\small
\setlength{\tabcolsep}{6pt}
\renewcommand{\arraystretch}{1}

\begin{tabularx}{\linewidth}{@{}p{0.14\linewidth}Y@{}}
\toprule
\textbf{Prompt}
& You are an expert in generating instance goal navigation instruction.\\
& Rules to generate the instruction \\
& 1. You will be given an specific object. \\
& 2. You will also be given the object's category, room, some attributes information, nearby objects and a detail description of the object. \\
& 3. You should try to generate a goal navigation instruction of the given object using all the given information \\
& 4. The instruction should be concise and include all the given information at the same time. \\
& 5. The format of the instruction can be diversity but has the similar meaning to ``Find the object.'' \\
& 6. You only need to output the final instruction, don't be verbose.\\ \\
& Now your input is: \\
& Object Category: \{category\} \\
& Room: \{room\} \\
& Attributes Information: \{attributes\_infos\} \\
& Nearby Objects: \{nearby\_objects\} \\
& Detail Description: \{description\}
\\
\bottomrule
\end{tabularx}
\end{table}

\FloatBarrier
\subsection{Training Trajectories collection}
\begin{figure}[htbp!]
\centering
\includegraphics[width=\linewidth]{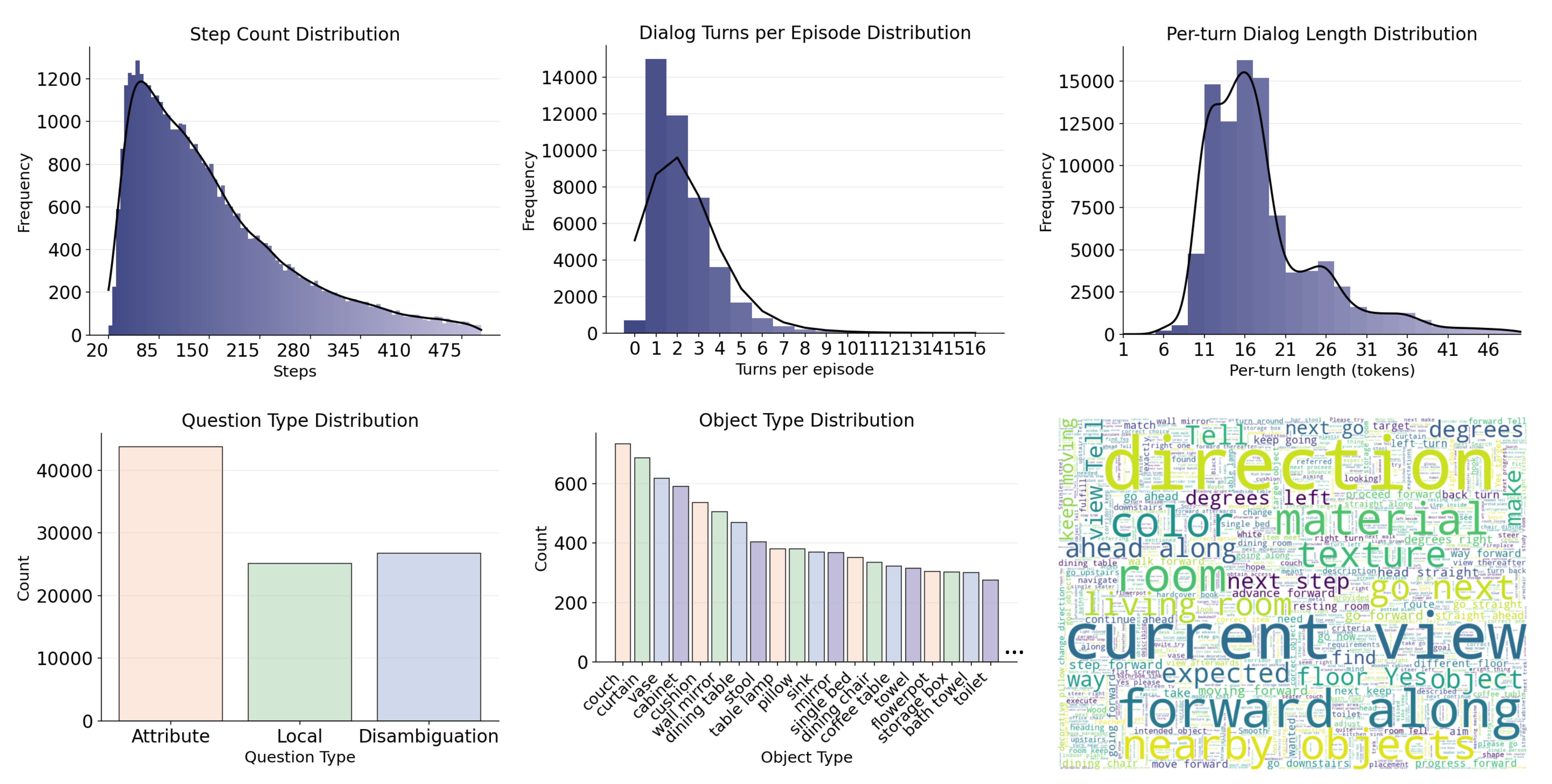}
\caption{\textbf{Statistics of the VL--LN training set.} The first row shows the distributions of step count, dialog turns per episode, and per-turn dialog length. The second row shows the distributions of question types and target object types, along with a word cloud of the dialog responses.}
\label{fig:dataset_statistics}
\end{figure}

\suppressfloats[t]
The frontier-based exploration (FBE) agent is equipped with a ground-truth object detector. At each selection, it chooses the next frontier in a stochastic yet goal-aware manner: with 90\% probability, it picks the frontier closest to the previously visited one to promote local, coherent exploration; with the remaining 10\% probability, it selects the frontier with the smallest geodesic distance to the target instance. This strategy balances broad coverage with occasional goal-directed guidance, improving search efficiency. The detector runs continuously, and once the target is observed, the agent navigates directly to it, yielding trajectories with purposeful and structured exploration. The pseudocode is provided in \cref{code: fbe}. Moreover, the statistics of the collected trajectories can be found in~\cref{fig:dataset_statistics}.

\begin{algorithm}[htbp!]
\caption{Frontier-Based Exploration with Ground-Truth Detector}
\label{code: fbe}
\KwIn{occupancy map $\mathcal{M}_o$, explored map $\mathcal{M}_e$, start pose $s_0$, target instance $\hat{o}_c$, max steps $T$}
\KwOut{Exploratory trajectory $\mathrm{traj}$}

$\mathrm{traj}\leftarrow[\ ]$, $s\leftarrow s_0$, $f_{\mathrm{prev}}\leftarrow \varnothing$\;

\For{$t\leftarrow 1$ \KwTo $T$}{
    \If(\tcp*[f]{detector runs continuously}){$\textsc{TargetSeen}(s,\hat{o}_c)$}{
        $\mathrm{traj}\leftarrow \mathrm{traj}\ \Vert\ \textsc{FollowShortestPath}(s,\hat{o}_c)$\;
        \Return{$\mathrm{traj}$}\;
    }

    $\mathcal{F}\leftarrow \textsc{GetFrontiers}(\mathcal{M}_o,\mathcal{M}_e)$\;
    \If{$|\mathcal{F}|=0$}{\Return{$\mathrm{None}$}\;}

    Draw $u\sim\mathcal{U}(0,1)$\;
    \eIf{$u<0.9$}{
        $g \leftarrow \arg\min_{f\in\mathcal{F}}\textsc{GeoDist}(f,f_{\mathrm{prev}})$\;
    }{
        $g \leftarrow \arg\min_{f\in\mathcal{F}}\textsc{GeoDist}(f,\hat{o}_c)$\;
    }
    $f_{\mathrm{prev}}\leftarrow g$\;

    $\Delta\mathrm{traj}, s, (\mathcal{M}_o,\mathcal{M}_e)\leftarrow \textsc{NavigateAndUpdate}(s,g,\mathcal{M}_o,\mathcal{M}_e)$\;
    $\mathrm{traj}\leftarrow \mathrm{traj}\ \Vert\ \Delta\mathrm{traj}$\;
}
\Return{$\mathrm{None}$}\;

\BlankLine
\textbf{Subroutines:}\;
\textsc{NavigateAndUpdate} follows the shortest path to $g$ while updating maps, logging states and actions.\;
\textsc{TargetSeen} returns true once the target is observed.\;
\end{algorithm}

\FloatBarrier
\subsection{Training Implementation}
\suppressfloats[t]

\noindent\textbf{Implementation Details.}
Our learning-based navigation model VLLN is initialized from \textit{Qwen3-VL-4B-Instruct}~\cite{qwen3technicalreport} and trained following the InternVLA-N1 training recipe~\cite{internvla-n1}. Training data consists of trajectories automatically collected by our pipeline from three tasks: ObjectNav (20{,}833 trajectories), IGN (17{,}224 trajectories), and IIGN (11{,}661 trajectories). Unless otherwise specified, VLLN is trained jointly on all three datasets. To analyze the contribution of different training sources, we further train three ablated variants, namely VLLN-O, VLLN-I, and VLLN-II, using only ObjectNav, IGN, or IIGN data, respectively.

\noindent\textbf{Training Setup.}
All four learned baselines were trained on the same GPU cluster with 32 NVIDIA A800 GPUs. Each training run took approximately 55 hours, corresponding to around 1{,}760 GPU-hours per model. Training was conducted in a distributed manner using DeepSpeed with BF16 mixed precision. We jointly fine-tuned the vision encoder, multimodal projector, and language model. We used 4 dataloader workers per process.

\noindent\textbf{Hyper-parameters.}
Unless otherwise specified, each model was trained for 5 epochs. We used a cosine learning-rate scheduler with a warmup ratio of 0.03, no weight decay, and gradient clipping with a maximum norm of 1.0. The per-device training batch size was set to 2, with 1 gradient accumulation step. The learning rate for the language model was set to \(2\times10^{-5}\), while the vision tower used a separate learning rate of \(5\times10^{-6}\). Training was conducted with BF16 mixed precision and gradient checkpointing, and we used 4 dataloader workers per process.

\noindent\textbf{Training Prompt.}
For supervised fine-tuning, each training sample is constructed as a multi-image, multi-turn vision-language conversation. The prompt begins with an instruction that defines the agent role, the target object category, the available action space, and the stopping criterion. Specifically, the model is told that it acts as an autonomous navigation assistant that may either \emph{move} or \emph{talk} to an oracle. If it chooses to talk, it should generate a natural-language query to help identify the target instance or improve navigation. If it chooses to move, it can either output a turn action or predict a pixel goal on the floor-view image. The episode terminates when the model outputs \texttt{STOP}.

The visual input in each prompt contains three parts. First, we provide a sequence of historical egocentric observations to summarize the recently visited context. Second, we include the previous look-down image, which is the floor-view image on which the agent previously predicted a waypoint. Third, we append the current egocentric observation, which represents the agent's latest visual state for high-level decision making. Unless otherwise noted, all historical observations and the current egocentric image are resized to \mbox{$384\times384$}, while the look-down image is kept at \mbox{$640\times480$}. This asymmetric design preserves more spatial detail in the floor-view image, which is directly used for waypoint localization.

For dialog-enabled samples, the prompt additionally includes the previous dialog history between the navigator and the oracle. In this way, the model conditions its next action not only on current visual observations, but also on past interaction history and previously acquired semantic evidence.

The supervision target follows our two-stage action formulation. For movement actions, the model first predicts a discrete control token, such as \texttt{<move>↓}, \texttt{<move>←}, or \texttt{<move>→}. After the tilt-down action, a follow-up prompt presents the corresponding look-down image and asks the model to localize the goal position on the floor, supervised by a pair of image coordinates. For dialog actions, the model directly generates a natural-language utterance to query the oracle. This unified prompt format allows VLLN to learn both embodied action prediction and task-oriented dialog generation within a single autoregressive framework.

\noindent\textbf{Simplified Prompt Examples.}
We provide three simplified examples to illustrate the training prompt format used for supervised fine-tuning.

\begin{quote}
\small
\raggedright
\textbf{Example 1 (move-only).}

\texttt{\_Human\_: You are an autonomous navigation assistant. Your task is to search for the bathtub. You may either move or talk to an oracle. If you choose to move, first output TILT DOWN (↓), and then predict the next waypoint coordinates on the floor-view image. If the target is reached, output STOP. These are your historical observations: <image> <image> <image> <image>. Your previous look-down image is <image>. In front of you is <image>.}

\texttt{\_Assistant\_: <move>↓}

\texttt{\_Human\_: You can see <image>. Locate the goal location on the floor in this image.}

\texttt{\_Assistant\_: <move>371 516}
\end{quote}

\begin{quote}
\small
\raggedright
\textbf{Example 2 (dialog then move).}

\texttt{\_Human\_: You are an autonomous navigation assistant. Your task is to search for the plant. You may either move or talk to an oracle. These are your historical observations: <image> <image> <image>. Your previous look-down image is <image>. In front of you is <image>.}

\texttt{\_Assistant\_: <talk>What is the color of the target?}

\texttt{\_Human\_: Green.}

\texttt{\_Assistant\_: <move>↓}

\texttt{\_Human\_: There is <image>. Locate the goal location on the floor in this image.}

\texttt{\_Assistant\_: <move>446 629}
\end{quote}

\begin{quote}
\small
\raggedright
\textbf{Example 3 (with dialog history).}

\texttt{\_Human\_: You are an autonomous navigation assistant. Your task is to search for the bowl. You may either move or talk to an oracle. These are your historical observations: <image> <image> <image> }
\texttt{<dialog> <|navigator|> Is this the bowl you asked for? <|oracle|> Correct.}
\texttt{\\</dialog>. Your previous look-down image is <image>. You can see \\<image>.}

\texttt{\_Assistant\_: <move>→→→→}
\end{quote}

\noindent In all examples, the historical observations and the current egocentric image are resized to \mbox{$384\times384$}, while the look-down image is kept at \mbox{$640\times480$}. Example~1 shows a move-only case, Example~2 shows how the model first queries the oracle and then continues navigation, and Example~3 shows how previous dialog history is incorporated into subsequent action prediction.

\FloatBarrier
\subsection{Real-world Implementation}
\suppressfloats[t]
\noindent\textbf{Real-world Experimental Details.}
We evaluate VLLN in a real-world setup using a quadruped robot (Unitree Go2) equipped with an Intel RealSense D455 RGB-D camera mounted at a height of 70\,cm with a \(15^\circ\) downward tilt. The full model runs on a remote server with an RTX 4090 GPU and occupies approximately 20\,GB of memory. During deployment, the robot streams synchronized RGB-D observations to the remote server for asynchronous inference.

Given a goal-oriented instruction, VLLN predicts either a motion action or a dialog action. For motion, it outputs either a turn command or a pixel goal on the floor-view image. If a pixel goal is predicted, we use the depth observation to project it into a 3D point goal, which is treated as the desired target position and passed to a PID-based local motion planner. The planner then converts the predicted action into executable linear and angular velocities. If the predicted action is a turn command, it is directly executed by the local motion planner. If VLLN chooses to ask a question, the robot sends the query to a human oracle and resumes navigation after receiving the reply.

\FloatBarrier
\section{Additional Experiment Results}
\subsection{Gap between Objectnav and IGN}
\begin{figure}[tb]
    \centering
    \includegraphics[width=0.55\linewidth]{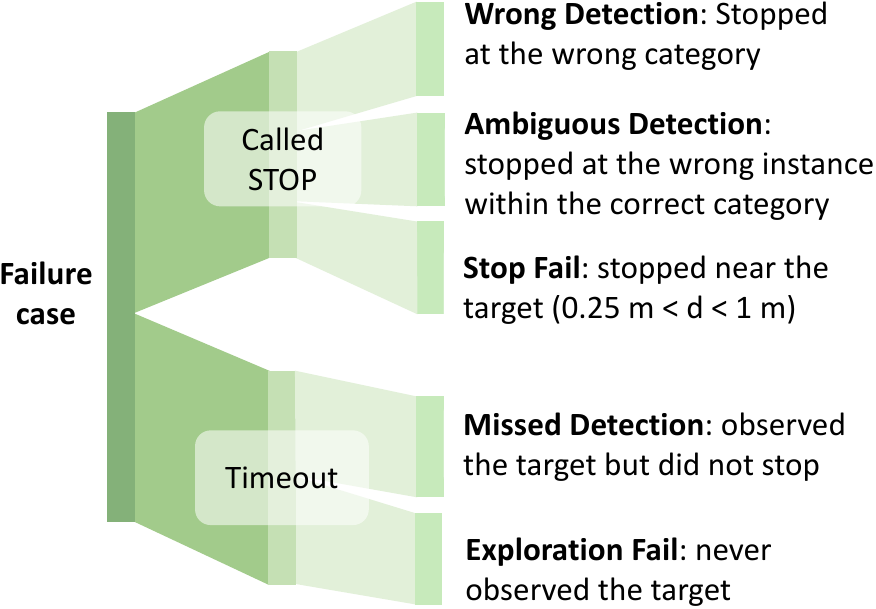}
    \caption{Failure taxonomy used in our error analysis.}
    \label{fig:supp_failure_analysis}
\end{figure}

\paragraph{Failure Categories.}
To better understand failure modes, we group failures into three categories, as shown in \cref{fig:supp_failure_analysis}: \textit{Detection Failure}, \textit{Exploration Failure}, and \textit{Stop Failure}. We further divide \textit{Detection Failure} into \textit{Missed Detection}, \textit{Wrong Detection}, and \textit{Ambiguous Detection}. \textit{Missed Detection} occurs when the agent observes the target but fails to recognize it, and continues exploring until the step budget is exhausted. \textit{Wrong Detection} refers to cases where the agent stops at an object from the wrong category. \textit{Ambiguous Detection} refers to cases where the agent stops at a same-category distractor rather than the true target. \textit{Exploration Failure} occurs when the agent never observes the target before reaching the maximum step limit. \textit{Stop Failure} occurs when the agent stops near the target (within 1\,m) but remains outside the stopping threshold (0.25\,m).

\begin{figure}[hptb!]
\centering
\includegraphics[width=0.9\linewidth]{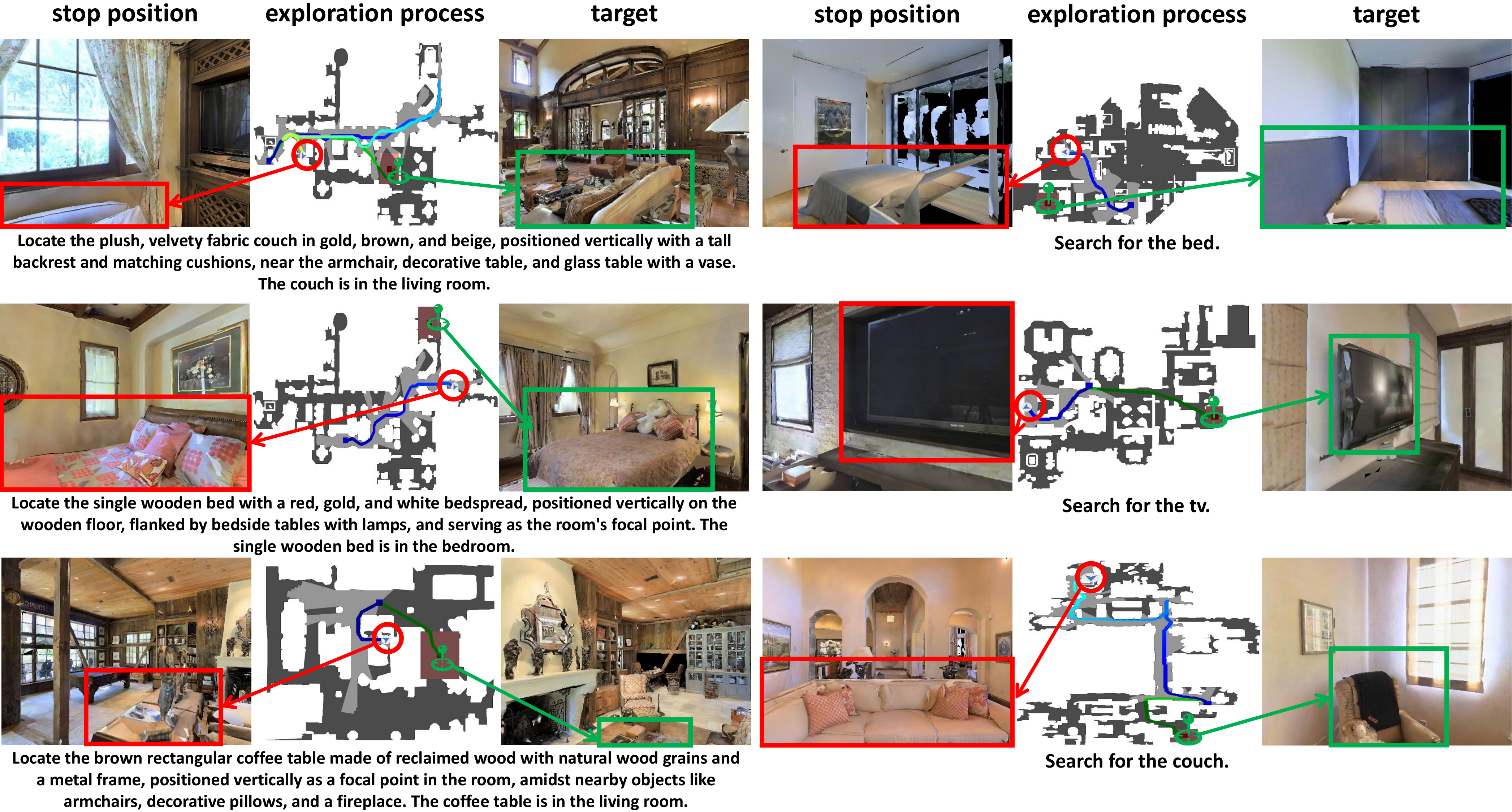}
\caption{\textbf{Visualization of ambiguity detection failure cases in IGN and IIGN.} For each row, the left side shows the IGN cases, while the right side shows the IIGN cases. Red boxes mark the agent’s stop location, green boxes mark the true target, and the middle maps show the exploration trajectories.}
\label{fig:AD_FAILURE}
\end{figure}

As shown in the main paper, the largest performance gap between ObjectNav and IGN comes from \textit{Ambiguous Detection}, which accounts for nearly 30\% of all evaluation episodes. Here we present several examples of \textit{Ambiguous Detection} failures (\cref{fig:AD_FAILURE}). The left examples are from IGN, while the right examples are from IIGN. In these cases, the agent stops at an object that belongs to the same category as the target but is not the correct instance. We attribute this failure mainly to insufficient attribute--image alignment, which prevents the agent from accurately grounding the complete instance description to the object with the corresponding attributes in the scene.

\FloatBarrier
\subsection{Human Failure Analysis}
\suppressfloats[t]
\begin{figure}[!ht]
\centering
\includegraphics[clip,width=0.85\linewidth]{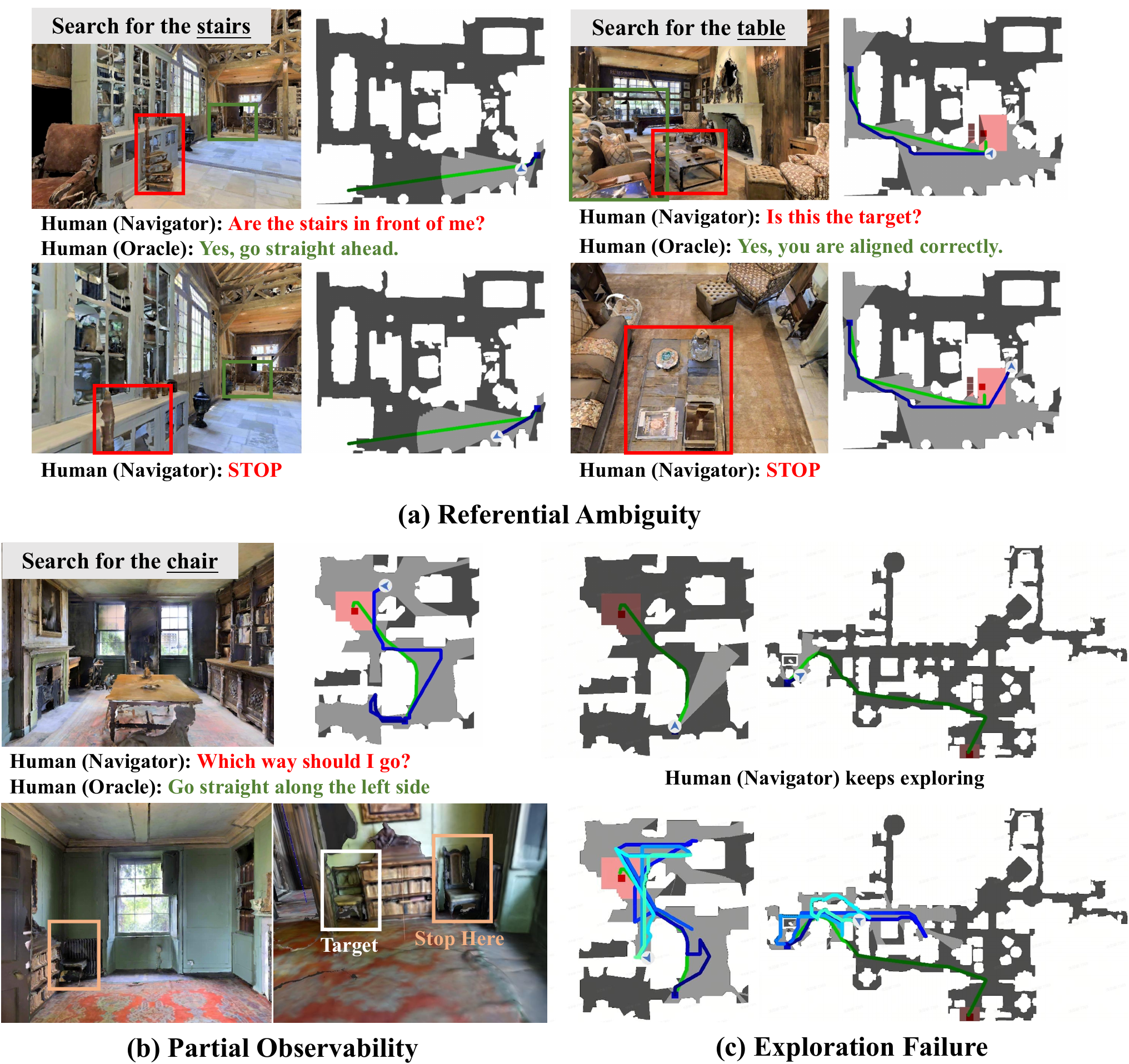}
    \caption{\textbf{Failure cases.} Green curves denote the geodesic shortest paths; blue curves are the navigator’s exploration trajectories; red shaded regions indicate the success zone around the target. (a) \emph{Referential ambiguity:} within the same view, the navigator and the NPC refer to different instances, causing the navigator to stop at a wrong instance. (b) \emph{Partial observability:} the navigator only observes a single candidate in the room and stops without disambiguating. (c) \emph{Exploration failure:} despite continued interaction, the human navigator never finds the target.}
    \label{fig:human_failure}
\end{figure}

As shown in Fig.~\ref{fig:human_failure}, two factors dominate in the human failure cases: (i) \emph{referential ambiguity}, where an expression such as ``the stairs'' does not uniquely identify the intended instance when multiple objects of the same category are visible, leading to incorrect grounding; and (ii) \emph{partial observability}, where the target is partially occluded, causing the navigator to assume a single candidate and commit without further disambiguation. These findings highlight an important direction for future IIGN research: enabling agents to localize and distinguish specific instances in the scene while actively asking clarification questions, since multiple candidate instances often appear in the same view. In the \emph{Human-GPT} setting, we observe an additional failure mode, \emph{exploration failure}, where the human navigator fails to finish the task before reaching the step limit. This usually occurs when the target is difficult to locate or the scene is large.

\FloatBarrier
\subsection{Question-Type Ablation}
\label{sec:question_type_ablation}

\begin{table}[htbp!]
\centering
\footnotesize
\caption{Question-type ablation on IIGN.}
\label{tab:qtype_ablation}
\setlength{\tabcolsep}{4pt}
\renewcommand{\arraystretch}{1.15}
\begin{tabular}{l c c c c c}
\toprule
\multicolumn{1}{l}{Allowed Question Types} &
\multicolumn{1}{c}{SR\,$\uparrow$} &
\multicolumn{1}{c}{SPL\,$\uparrow$} &
\multicolumn{1}{c}{OS\,$\uparrow$} &
\multicolumn{1}{c}{NE\,$\downarrow$} &
\multicolumn{1}{c}{$\Delta$SR $\uparrow$} \\
\midrule
-                         & 19.6 & 12.32  & 28.6 & 9.70 & 0.0 \\
Attr.                     & 20.8 & 12.66  & 34.6 & 9.27 & 1.2 \\
Disamb.                   & 21.8 & 13.32  & 32.8 & 9.31 & 2.2 \\
Route                     & 22.0 & 13.93  & 33.8 & 8.95 & 2.4 \\
Attr.+Route               & 23.0 & 14.72  & 35.6 & 8.59 & 3.4 \\
Attr.+Disamb.             & 21.2 & 12.76  & 33.4 & 8.95 & 1.6 \\
Route+Disamb.             & 21.0 & 12.56  & 33.8 & 8.48 & 1.4 \\
Attr.+Disamb. + Route     & 25.4 & 16.18  & 35.2 & 8.61 & 5.8 \\
\bottomrule
\end{tabular}
\end{table}
\FloatBarrier

\noindent We evaluate each question type, each pair, and all three types while keeping all other settings fixed. As shown in~\cref{tab:qtype_ablation}, attribute (Attr.), disambiguation (Disamb.), and route questions independently improve SR by 1.2, 2.2, and 2.4 points, respectively. Although route questions provide the largest single-type gain, they account for only $41.38\%$ of the full improvement, showing that the overall gain is not driven by route guidance alone. Among the two-type settings, Attr.+Route achieves the largest gain (+3.4). These results demonstrate the distinct benefits of the three question types.

\FloatBarrier
\subsection{Oracle-Backbone Ablation}
\label{sec:oracle_backbone_ablation}

\begin{table}[htbp!]
\centering
\footnotesize
\caption{Oracle-backbone ablation on IIGN.}
\label{tab:oracle_backbone_ablation}
\setlength{\tabcolsep}{4pt}
\renewcommand{\arraystretch}{1.15}
\begin{tabular}{l c c c c}
\toprule
\multicolumn{1}{l}{Oracle LLM} &
\multicolumn{1}{c}{SR\,$\uparrow$} &
\multicolumn{1}{c}{SPL\,$\uparrow$} &
\multicolumn{1}{c}{OS\,$\uparrow$} &
\multicolumn{1}{c}{NE\,$\downarrow$} \\
\midrule
GPT-4o                     & 25.4 & 16.18  & 35.2 & 8.61 \\
Qwen3-VL-4B-Instruct       & 24.8 & 15.36  & 38.2 & 8.50 \\
Kimi2.5                    & 25.2 & 16.28  & 36.8 & 8.40 \\
DeepSeek-R1                & 25.0 & 16.19  & 37.6 & 8.64 \\
\bottomrule
\end{tabular}
\end{table}
\FloatBarrier

To assess evaluation sensitivity to the oracle backbone, we replace GPT-4o with three alternatives while keeping all other settings fixed. As shown in~\cref{tab:oracle_backbone_ablation}, SR varies by only 0.6 points (24.8--25.4), indicating that VL-LN Bench is robust to the oracle choice and reproducible with open-source or lower-cost models.

\FloatBarrier
\subsection{Dialog-Turn Budget Ablation} 

\begin{table}[htbp!]
\centering
\footnotesize
\caption{Results of dialog-turn ablation.}
\label{tab:dialog_limits}
\setlength{\tabcolsep}{4pt}
\renewcommand{\arraystretch}{1.15}
\begin{tabular}{l c c c c c}
\toprule
\multicolumn{1}{l}{Turn limit} &
\multicolumn{1}{c}{SR\,$\uparrow$} &
\multicolumn{1}{c}{SPL\,$\uparrow$} &
\multicolumn{1}{c}{OS\,$\uparrow$} &
\multicolumn{1}{c}{NE\,$\downarrow$} &
\multicolumn{1}{c}{Avg.\ turns} \\
\midrule
0            & 19.6 & 12.32  & 28.6 & 9.70 & 0.00 \\
1            & 22.0 & 13.67  & 32.6 & 9.01 & 1.00 \\
2            & 23.0 & 15.15  & 31.4 & 9.14 & 1.71 \\
3            & 23.0 & 15.19  & 33.0 & 8.74 & 1.95 \\
4            & 25.2 & 15.80  & 34.0 & 8.68 & 1.98 \\
$\infty$ (5) & \textbf{25.4} & \textbf{16.18} & \textbf{35.2} & \textbf{8.61} & 1.98 \\
\bottomrule
\end{tabular}
\end{table}
\FloatBarrier

We evaluate VLLN under different dialog-turn limits to study the effect of dialog budget. As shown in~\cref{tab:dialog_limits}, SR and SPL generally improve as the budget increases, demonstrating the benefit of interaction. The largest improvement occurs when increasing the budget from 0 to 1 turn, likely because the first question is often an attribute question that helps eliminate distractors and narrow down candidate instances. Meanwhile, the agent remains query-efficient: even with larger budgets, it asks fewer than two questions on average (1.71--1.98). This trend is consistent with the training distribution, where the average number of dialog turns is around 1.9, as shown in Fig.~\ref{fig:dataset_statistics}(b), suggesting that query efficiency and dialog frequency may both be shaped by the training data.

\FloatBarrier
\section{Limitation and Future Work}
\subsection{Ambiguity Detection}
A major limitation revealed by our analysis is weak image--attribute alignment. A promising direction is to explicitly strengthen ambiguity detection during training. One simple strategy is to introduce hard negatives, \eg, same-category objects with different colors, materials, or relative placements, so that the agent learns to discriminate between visually similar distractors. 
\subsection{Learning Better Dialog Policies}
The IIGN task poses new challenges for dialog policy learning. Rather than relying primarily on supervised imitation of collected questioning trajectories, future agents could learn question-asking policies through reinforcement learning, with rewards designed to encourage ambiguity reduction, efficient exploration, and task completion with minimal turns. 

\subsection{Toward a More Helpful Oracle}
Although the current oracle is already reliable enough for evaluation, it is not yet fully aligned with human assistance. \emph{Human--GPT} achieves performance close to \emph{Human--Human} (91\% vs.\ 93\% SR), supporting the validity of the benchmark, but requires substantially more dialog turns on average (9.72 vs.\ 2.04). We attribute this gap to two main factors. First, some user queries fall outside the oracle's current knowledge or response schema. Second, users interacting with AI systems often seek additional confirmation even after the target has effectively been identified. These observations suggest that, although the current oracle is dependable, it is not yet as helpful or natural as a human assistant. A more helpful oracle could therefore go beyond responding within a fixed schema. Future versions may provide richer pragmatic feedback, such as uncertainty-aware responses, candidate comparison, proactive clarification, or adaptive route guidance conditioned on the agent's progress and failure history.

\end{document}